\documentclass[times,twocolumn,final]{elsarticle}

\usepackage[
  left=2.35cm,
  right=2.35cm,
  top=1.8cm,
  bottom=2.0cm,
  columnsep=0.6cm
]{geometry}

\usepackage{graphicx}
\usepackage{booktabs}
\usepackage{tabularx}
\usepackage{array}
\usepackage{multirow}
\usepackage{makecell}
\usepackage{caption}
\usepackage{framed}
\newcolumntype{Y}{>{\centering\arraybackslash}X}

\usepackage{amsmath}
\usepackage{amssymb}
\usepackage{mathtools}
\usepackage{latexsym}
\usepackage{bm}

\usepackage[ruled,vlined]{algorithm2e}
\SetKwComment{Comment}{$\triangleright$\ }{}

\usepackage{xcolor}
\usepackage{url}
\usepackage{pifont}

\usepackage[hidelinks]{hyperref}

\definecolor{newcolor}{rgb}{.8,.349,.1}

\begin{document}

\begin{frontmatter}

\title{Less Is More: Training-Free Acceleration Framework of 3D Diffusion Models for Low-Count PET Denoising via Global–Local Trajectory Reduction}

\author[1,3]{Yuhan Liu}
\author[1]{Scott M. Leonard}
\author[1]{Marlee Crews}
\author[1]{Muhannad Fadhel}
\author[1]{Jinkui Hao}
\author[1]{Tianqi Chen}
\author[1]{Ryan J. Avery}
\author[1]{Bo Zhou\corref{cor1}}
\cortext[cor1]{Bo Zhou: \textit{bo.zhou@northwestern.edu}} 
\address[1]{Department of Radiology, Northwestern University, Chicago, IL, USA}
\address[3]{Department of Biomedical Engineering, Hefei University of Technology, Hefei, Anhui, China}
\tnotetext[dagger]{Part of the data used in preparation of this article was obtained from the University of Bern,
Dept. of Nuclear Medicine and School of Medicine, Ruijin Hospital. As such, the investigators
contributed to the design and implementation of DATA and/or provided data, but did not participate
in the analysis or writing of this report. A complete listing of investigators can be found at:
\url{https://udpet-challenge.github.io/}.
}

\begin{abstract}
Accurate quantification and uptake measurement in Positron Emission Tomography (PET) are critical for assessing disease progression and supporting clinical decision-making. While high-count PET provides reliable image quality, the associated radiation dose and prolonged acquisition remain significant clinical concerns, motivating the adoption of low-count protocols.
Diffusion-model-based methods have demonstrated strong potential for restoring low-count PET to near high-count quality, but their iterative sampling procedure becomes prohibitively expensive when applied to high-resolution 3D PET volumes, introducing substantial inference latency that limits practical clinical deployment.
To address these challenges, we propose a training-free Global-Local Skipping Strategy that accelerates diffusion model-based 3D PET denoising while simultaneously improving reconstruction quality. The proposed method is plug-and-play and directly applicable to pre-trained diffusion models without retraining or architectural modification. 
Specifically, we introduce: (i) a global denoising step skipping strategy that initializes the reverse diffusion process from an intermediate denoising step using a noise-consistent transformation of the low-count input, substantially reducing the number of required denoising steps; and (ii) a local feature reuse shortcut that reuses slowly-varying high-level U-Net features across neighboring denoising steps, further reducing per-step computation while preserving image fidelity.
We evaluate the proposed approach on multiple PET tracers from in-house and public datasets, including $^{18}$F-FDG PET, $^{68}$Ga-DOTATATE PET, and $^{18}$F-PSMA PET, demonstrating consistent acceleration of over an order of magnitude alongside improved or comparable reconstruction performance relative to the full-step baseline. Blinded reader studies further confirm enhanced clinical confidence and perceived diagnostic quality.
{Code is available at: \href{https://github.com/Advanced-AI-in-Medicine-and-Physics-Lab/LIM.git}{\nolinkurl{https://github.com/Advanced-AI-in-Medicine-and-Physics-Lab/LIM.git}}}

\end{abstract}


\end{frontmatter}

\section{Introduction}
Positron Emission Tomography (PET) is a cornerstone imaging modality in oncology, cardiology, and neurology, enabling quantitative assessment of radiotracer uptake for disease diagnosis, treatment planning, and therapy monitoring. Reliable quantification typically requires high-count acquisitions to ensure adequate count statistics and image quality. However, higher count levels necessitate increased injected activity or prolonged acquisition times, leading to greater radiation exposure and reduced practicality for longitudinal imaging, vulnerable patient populations, and population-scale screening. Low-count PET provides a clinically attractive alternative by reducing administered dose or shortening scan duration, but suffers from elevated noise, degraded image quality, and compromised quantitative accuracy, which can adversely affect lesion detection, segmentation, and standardized uptake value (SUV) measurements.

To mitigate these limitations, numerous deep learning-based PET denoising approaches have been proposed. Among them, denoising diffusion probabilistic models (DDPMs)~\citep{ho_denoising_2020} have recently achieved state-of-the-art performance across scanners, tracers, and count levels~\citep{yu2025robust}, owing to their powerful generative priors and ability to progressively refine degraded images through iterative denoising. Nevertheless, the exceptional restoration capability of diffusion models comes at a substantial computational cost. Modern diffusion models typically require hundreds to thousands of reverse denoising steps during inference, and this burden becomes particularly severe for high-resolution 3D medical imaging. Recent studies have reported that denoising a single whole-body PET scan using a 3D diffusion model may require tens of minutes to several hours of GPU computation~\citep{yu2025robust}, creating a major obstacle for clinical deployment. Consequently, many existing approaches resort to reduced-dimensional approximations, such as 2.5D diffusion models~\citep{2.5DMulti-ViewPET}, sacrificing full volumetric modeling in exchange for computational feasibility.

Considerable effort has been devoted to accelerating diffusion models through scheduler redesign~\citep{song2022denoisingdiffusionimplicitmodels,DPMSolver,DPM++}, knowledge distillation~\citep{ProgressiveDistillation}, learned fast samplers~\citep{FastDDPMMedicalImageGeneration}, and architectural modifications~\citep{2.5DMulti-ViewPET}. While these approaches have demonstrated promising acceleration, many require retraining, fine-tuning, or modifications to the underlying network architecture. Moreover, most recent acceleration techniques have been developed for large-scale natural-image generation and are closely coupled with Diffusion Transformers (DiTs)~\citep{DiffusionTransformers}, limiting their applicability to the U-Net-based diffusion architectures that remain dominant in medical image restoration.

More fundamentally, existing acceleration methods largely overlook an important distinction between image generation and image restoration. In natural-image generation, the reverse diffusion process must progressively synthesize image content from pure Gaussian noise. In contrast, medical image restoration is inherently a conditional problem: the degraded input already contains substantial anatomical and functional information about the target image. In low-count PET, despite elevated noise levels, the input image preserves most anatomical structures and coarse tracer uptake patterns of the corresponding high-count image. This observation raises an intriguing question: \emph{is traversing the full diffusion trajectory necessary when the desired image structure is already largely present in the conditional input?}

In this work, we demonstrate that the answer is often no. We show that diffusion-based PET denoising contains substantial redundancy at two complementary levels. First, the early portion of the reverse diffusion trajectory is largely redundant because the low-count PET image already provides a structurally informative initialization. Second, intermediate representations within the diffusion U-Net evolve smoothly across neighboring reverse denoising steps, resulting in considerable redundancy in network computation. These observations suggest that both the diffusion trajectory and the denoising network contain exploitable inefficiencies that can be removed without retraining the model.

Motivated by these insights, we propose \textit{Less Is More (LIM)}, a training-free and plug-and-play acceleration framework for 3D diffusion-based PET denoising. LIM introduces a unified \textit{Global--Local Skipping} strategy consisting of two complementary components. At the global level, we introduce a \textbf{Q-Sampling Trajectory Shortcut}, which initializes reverse diffusion from an intermediate noisy state derived from the low-count PET image, thereby bypassing a large portion of the reverse trajectory. At the local level, we introduce a \textbf{Feature-Reuse Network Shortcut}, which exploits the temporal smoothness of high-level U-Net representations by caching and reusing intermediate features across neighboring reverse denoising steps. Importantly, the proposed framework requires neither retraining nor architectural modification and can be directly applied to existing pre-trained diffusion models as an inference-time acceleration strategy.

We evaluate LIM on both public and in-house PET datasets spanning multiple tracers, including $^{18}$F-FDG, $^{68}$Ga-DOTATATE, and $^{18}$F-PSMA PET. Experimental results demonstrate acceleration factors exceeding one order of magnitude while consistently improving or maintaining reconstruction fidelity relative to the original diffusion process. Notably, the proposed method achieves superior quantitative performance despite substantially reducing computational cost, suggesting that removing redundant portions of the diffusion trajectory may also mitigate cumulative denoising errors. Blinded reader studies further demonstrate improved clinical confidence and lesion visibility. The main contributions of this work are summarized as follows:

\begin{itemize}
\setlength{\itemsep}{2pt}
\setlength{\parskip}{0pt}
\setlength{\parsep}{0pt}

\item \textbf{A training-free acceleration framework for 3D diffusion-based PET denoising.}
We propose a plug-and-play inference strategy that substantially accelerates pre-trained diffusion models without retraining, fine-tuning, or architectural modification.

\item \textbf{A unified Global--Local Skipping paradigm.}
We introduce complementary trajectory-level and network-level acceleration mechanisms that exploit redundancy in both the diffusion process and the denoising network.

\item \textbf{A restoration-specific perspective on diffusion acceleration.}
We provide empirical and theoretical evidence that conditional medical image restoration exhibits substantial trajectory and feature redundancy, fundamentally differing from unconditional image generation.

\item \textbf{Comprehensive multi-tracer validation.}
We validate the proposed framework across multiple PET tracers and datasets, demonstrating consistent improvements in reconstruction quality, computational efficiency, and blinded clinical evaluation.
\end{itemize}

\begin{figure*}[htb!]
    \centering
    \includegraphics[width=0.93\textwidth]{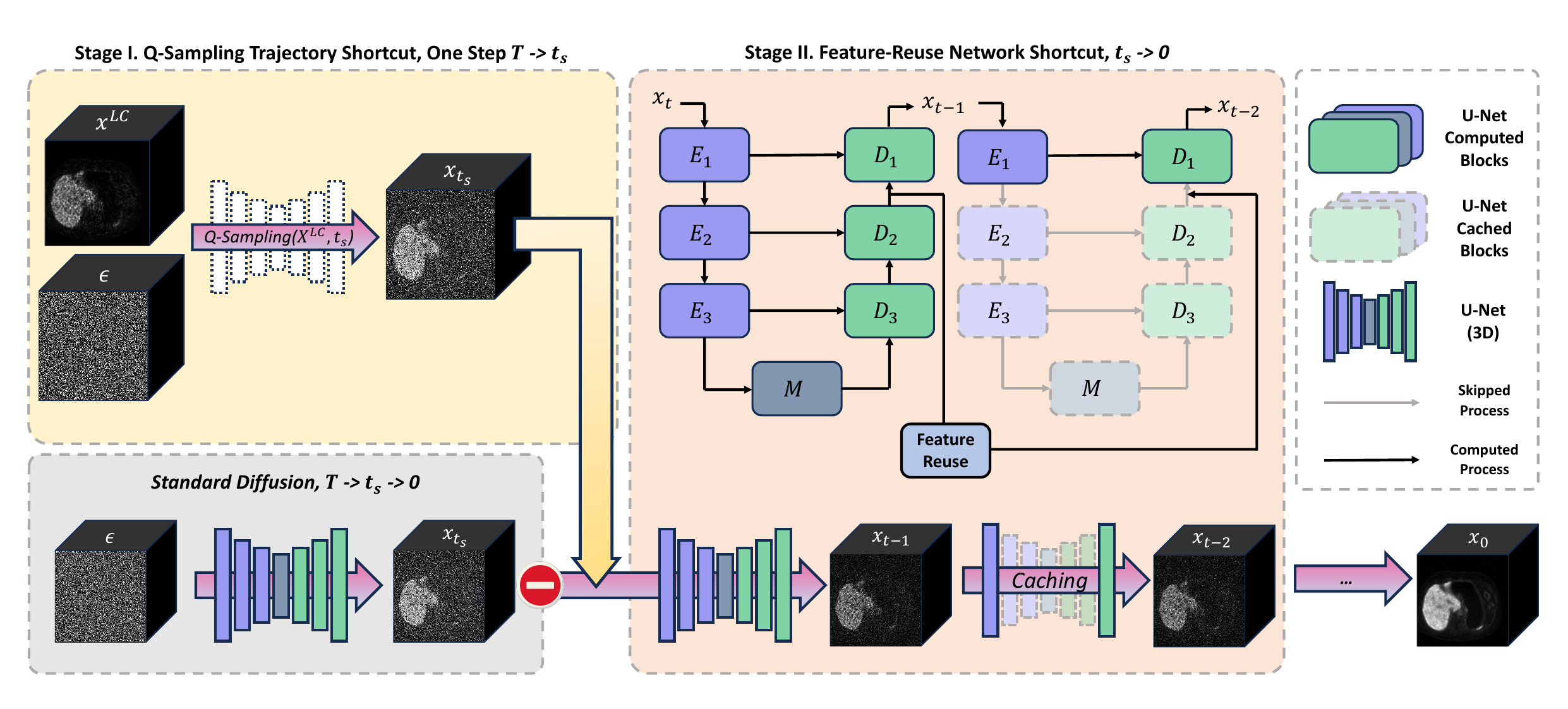}
    \vspace{-8pt}
    \caption{Overview of the proposed \textbf{Less Is More (LIM)} framework for training-free acceleration of diffusion-based PET denoising. The framework consists of two complementary acceleration stages. \textbf{Stage I: Q-Sampling Trajectory Shortcut} performs \emph{global trajectory reduction} by initializing the reverse diffusion process from an intermediate timestep $t_s$ instead of pure Gaussian noise at $T$, using a Q-sampled low-count PET image constructed from the observed low-count input and matched Gaussian perturbation. This initialization preserves image-specific anatomical and tracer uptake information while bypassing redundant early reverse denoising steps. \textbf{Stage II: Feature-Reuse Network Shortcut} performs \emph{local computation reduction} by caching and reusing intermediate U-Net representations across neighboring reverse denoising steps, avoiding repeated computation of temporally similar features. Together, the two stages reduce both the overall diffusion trajectory length and per-step computational cost, enabling substantially faster inference while maintaining reconstruction fidelity.}
    \label{Fig-Method}
\end{figure*}

\vspace{-0.2cm}
\section{Related Works}
\label{sec-relatedwork}

\subsection{Deep Learning for Low-Count PET Denoising}
\label{relatedworks-DL-PET-Denoising}

Low-count PET denoising has been extensively studied as a strategy to reduce radiation exposure while maintaining clinically reliable image quality and quantitative accuracy. Most existing approaches formulate the problem as supervised image-to-image restoration by learning a mapping from low-count PET to corresponding high-count PET images, typically generated through retrospective list-mode downsampling.

Early deep learning approaches were primarily based on convolutional neural networks (CNNs), particularly U-Net-style encoder--decoder architectures~\citep{UNet,UNetPETRecon,mazandarani2023unext}, which effectively capture multi-scale contextual information while preserving anatomical structures through skip connections. Subsequently, generative adversarial networks (GANs) and conditional GANs were introduced to improve perceptual realism and image sharpness~\citep{kaplan2019full,3D-cGAN-PET}, followed by CycleGAN- and Wasserstein GAN-based variants~\citep{Supervised-CycleGAN-FDGDenoising,PT-WGAN-LDPET-Denoising}. Although adversarial learning often improves visual quality, it may introduce hallucinated structures or unstable training behavior, which are undesirable in quantitative PET applications. To further improve restoration fidelity under extremely low-count settings, several studies incorporated complementary anatomical information from MRI, such as T1-weighted or multi-contrast images, as structural priors for PET restoration~\citep{chen2019ultra,xiang2017deep}.

More recently, diffusion models have emerged as the state-of-the-art paradigm for PET denoising and medical image restoration~\citep{yu2025robust,yu2024pet,DoseAwareDiffusionModels}. By iteratively refining noisy observations through a learned reverse diffusion process, diffusion models achieve superior restoration fidelity and improved preservation of anatomical and functional details compared with earlier CNN- and GAN-based methods. In particular, both fully volumetric 3D diffusion models and more computationally efficient 2.5D diffusion frameworks have demonstrated promising results across scanners, tracers, and dose levels~\citep{yu2025robust,2.5DMulti-ViewPET}. However, these gains come at a substantial computational cost. The iterative denoising process requires hundreds to thousands of network evaluations, resulting in significantly longer inference times and higher memory consumption than conventional restoration networks. This limitation is especially pronounced in high-resolution 3D PET imaging, where inference for a single whole-body scan may require tens of minutes to several hours~\citep{yu2025robust}, motivating the need for efficient diffusion acceleration strategies.

\subsection{Acceleration of Diffusion Models}

A large body of recent work has focused on accelerating diffusion models~\citep{LightWeightDiffusionModelSurvey}. Existing approaches can be broadly categorized into architecture-level acceleration, sampling acceleration, model compression, and inference-time optimization. At the architecture level, one practical strategy for volumetric medical imaging is to replace fully 3D diffusion models with 2.5D architectures~\citep{2.5DMulti-ViewPET,hu2024diffgepci,XIE2026104039}. By processing neighboring slices or orthogonal views instead of the entire volume, these methods substantially reduce memory and computational requirements while partially preserving volumetric consistency. However, 2.5D models inherently approximate long-range three-dimensional dependencies and may be suboptimal for whole-body PET imaging, where tracer biodistribution patterns and lesion continuity extend across large spatial regions. Another major direction focuses on accelerating the reverse diffusion trajectory. Representative examples include DDIM~\citep{song2022denoisingdiffusionimplicitmodels}, DPM-Solver~\citep{DPMSolver}, and DPM-Solver++~\citep{DPM++}, which reduce the number of reverse denoising steps through alternative numerical formulations. These methods achieve substantial acceleration without retraining and have become standard baselines in diffusion generation. However, they are primarily developed for image synthesis and latent-space diffusion models. Aggressive trajectory shortening may lead to instability or fidelity degradation in conditional restoration tasks, where preserving subtle anatomical structures and quantitative image characteristics is critical. A third category consists of model compression techniques, including distillation, pruning, and quantization. Progressive Distillation~\citep{ProgressiveDistillation}, for example, compresses long diffusion trajectories into a smaller number of reverse denoising steps by training a student network to mimic a pre-trained teacher. While effective, these methods typically require additional training and often introduce deployment complexity or task-specific adaptation.

Recent advances in efficient diffusion inference have been largely driven by latent diffusion models~\citep{LatentDiffusionModels} and Diffusion Transformers (DiTs)~\citep{DiffusionTransformers}. Within this context, several training-free acceleration methods have been proposed. FORA~\citep{FORApaper} exploits temporal redundancy by reusing block-level features across neighboring reverse denoising steps, while ToCa~\citep{ToCa} selectively updates informative tokens and reuses cached representations for less important tokens. Sparse attention methods such as Jenga~\citep{Jenga} further reduce computational cost by restricting attention computation to structured subsets of tokens. These methods demonstrate that significant redundancy exists within diffusion trajectories and intermediate representations.

Despite this progress, most existing acceleration techniques have been developed for natural-image or video generation~\citep{StableDiffusion,labs2025flux1kontextflowmatching,OpenSora,WAN} and rely on assumptions that do not directly translate to medical image restoration. First, they are often tightly coupled with latent diffusion frameworks or transformer-based architectures, whereas U-Net-based diffusion models remain the dominant backbone in medical imaging. Second, they are designed for content generation from Gaussian noise rather than conditional restoration from an observed image. In low-count PET denoising, the input image already contains substantial anatomical and functional information, suggesting that a different form of redundancy may exist within the diffusion trajectory. To our knowledge, this restoration-specific redundancy has not been systematically exploited for accelerating 3D diffusion-based PET denoising. The proposed method addresses this gap through a training-free Global--Local Skipping framework that leverages both trajectory-level and feature-level redundancy during inference.

\section{Methods}
\label{sec:method}

\begin{figure*}[htb!]
    \centering
    \includegraphics[width=0.99\textwidth]{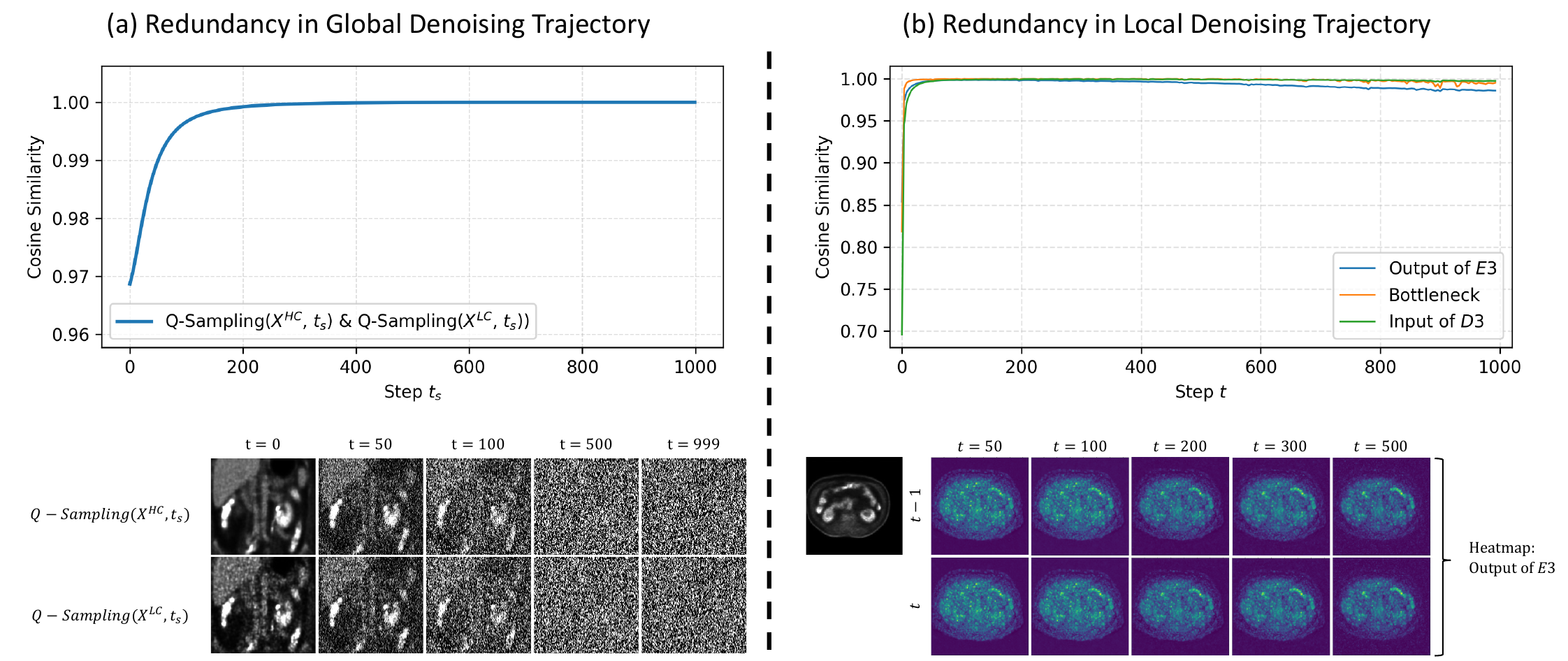}
    \vspace{-8pt}
    \caption{Redundancy Analysis in 3D Diffusion-based PET Denoising. \textbf{(a) Global redundancy: pairwise cosine similarity between q-sampled HC and LC PET states at the same denoising step.} Similarity remains consistently close to 1.0 across the entire diffusion trajectory (0.97 even in the original image domain), indicating that low-count PET preserves sufficient structural information to support intermediate initialization of the reverse process. \textbf{(b) Local redundancy: cosine similarity between features at adjacent reverse denoising steps across U-Net stages.} High temporal similarity across encoder, bottleneck, and decoder stages throughout the full trajectory ($t \in [0, 1000]$) demonstrates that intermediate representations evolve smoothly between consecutive reverse denoising steps, which supports effective feature reuse.}
\label{Fig-Intro}
\end{figure*}

The overall framework of LIM is illustrated in Fig.~\ref{Fig-Method} and is built upon a unified \textit{Global--Local Skipping} paradigm that exploits two complementary forms of redundancy in diffusion-based PET denoising, as characterized in Sec.~\ref{method:observation}. Specifically, we observe that redundancy exists both along the diffusion trajectory and within the denoising network itself. Based on these observations, we develop two training-free acceleration strategies. First, we introduce the \textbf{Q-Sampling Trajectory Shortcut} (Sec.~\ref{method:QTS}), a \emph{global} acceleration mechanism that leverages the strong structural correspondence between low-count and high-count PET images to initialize the reverse diffusion process from an intermediate timestep $t_s$, thereby bypassing a large portion of the redundant high-noise trajectory. Second, we introduce the \textbf{Feature-Reuse Network Shortcut} (Sec.~\ref{method:Cache}), a \emph{local} acceleration mechanism that exploits the smooth temporal evolution of high-level U-Net representations across neighboring reverse denoising steps, enabling cached features to be reused and reducing redundant network computation. Together, these two components accelerate inference at complementary levels: the global shortcut reduces the overall number of reverse denoising steps, while the local shortcut decreases the computational cost of each remaining denoising step. A complete algorithmic description of the proposed framework is provided in Algorithm~\ref{alg:fast_infer}.

\subsection{Preliminary: Conditional Denoising Diffusion Probabilistic Models}
\label{sec:ddpm_prelim}

Denoising Diffusion Probabilistic Models (DDPMs) are generative models that learn a data distribution by inverting a gradual noising process.
They define (i) a \emph{forward} Markov chain that progressively perturbs data into Gaussian noise, and (ii) a \emph{reverse-time} Markov chain parameterized by a neural network that iteratively removes noise.
In conditional image restoration tasks, the reverse process is further conditioned on an observed input $\bm{c}$.
In our PET denoising setting, $\bm{c}$ denotes the low-count PET image $x^{LC}$, and the model aims to recover the corresponding high-count PET image $x_0=x^{HC}$.

\paragraph{Forward diffusion}
Let $\bm{x}_0 \sim q(\bm{x}_0)$ denote a clean target image.
The forward diffusion process is a fixed Gaussian Markov chain:
\begin{align}
q(\bm{x}_t \mid \bm{x}_{t-1})
&= \mathcal{N}\!\left(\bm{x}_t;\ \sqrt{\alpha_t}\,\bm{x}_{t-1},\ (1-\alpha_t)\bm{I}\right),
\qquad t=1,\dots,T,
\label{eq:ddpm_forward}
\end{align}
where $\alpha_t = 1-\beta_t$ and $\{\beta_t\}_{t=1}^T$ is a predefined variance schedule with $\beta_t\in(0,1)$.
Define $\bar{\alpha}_t=\prod_{s=1}^t \alpha_s$.
By Gaussian composition, the marginal distribution admits a closed form:
\begin{align}
q(\bm{x}_t \mid \bm{x}_0)
&= \mathcal{N}\!\left(\bm{x}_t;\ \sqrt{\bar{\alpha}_t}\,\bm{x}_0,\ (1-\bar{\alpha}_t)\bm{I}\right),
\label{eq:ddpm_q_xt_x0}
\end{align}
which enables direct sampling via reparameterization:
\begin{align}
\bm{x}_t
&= \sqrt{\bar{\alpha}_t}\,\bm{x}_0 + \sqrt{1-\bar{\alpha}_t}\,\bm{\epsilon},
\qquad \bm{\epsilon}\sim\mathcal{N}(\bm{0},\bm{I}).
\label{eq:ddpm_reparam}
\end{align}

\paragraph{Conditional reverse process}
Given the condition image $\bm{c}$, the generative model starts from $\bm{x}_T\sim\mathcal{N}(\bm{0},\bm{I})$ and learns a conditional reverse-time Markov chain:
\begin{align}
p_\theta(\bm{x}_{t-1}\mid \bm{x}_t, \bm{c})
&= \mathcal{N}\!\left(\bm{x}_{t-1};\ \bm{\mu}_\theta(\bm{x}_t,t,\bm{c}),\ \bm{\Sigma}_t\right),
\qquad t=T,\dots,1,
\label{eq:ddpm_reverse}
\end{align}
where $\theta$ parameterizes a neural network and $\bm{\Sigma}_t$ is typically fixed as $\sigma_t^2\bm{I}$ for stable sampling.
A standard parameterization predicts the injected noise $\bm{\epsilon}_\theta(\bm{x}_t,t,\bm{c})$, yielding:
\begin{align}
\bm{\mu}_\theta(\bm{x}_t,t,\bm{c})
&=\frac{1}{\sqrt{\alpha_t}}
\left(
\bm{x}_t - \frac{\beta_t}{\sqrt{1-\bar{\alpha}_t}}\ \bm{\epsilon}_\theta(\bm{x}_t,t,\bm{c})
\right).
\label{eq:ddpm_mu_eps}
\end{align}

\paragraph{Training objective}
In practice, the model is commonly trained to predict the injected noise under the condition $\bm{c}$:
\begin{align}
\mathcal{L}_{\text{simple}}(\theta)
&=
\mathbb{E}_{t \sim \mathcal{U}\{1,\dots,T\},\ \bm{x}_0,\bm{c},\ \bm{\epsilon}\sim\mathcal{N}(\bm{0},\bm{I})}
\left[
\left\|
\bm{\epsilon} - \bm{\epsilon}_\theta(\bm{x}_t,t,\bm{c})
\right\|_2^2
\right],
\label{eq:ddpm_loss_simple}
\end{align}
where $\bm{x}_t = \sqrt{\bar{\alpha}_t}\,\bm{x}_0 + \sqrt{1-\bar{\alpha}_t}\,\bm{\epsilon}$.

\paragraph{Sampling}
At inference time, conditioned on $\bm{c}$, the model iteratively samples:
\begin{align}
\bm{x}_{t-1}
&= \bm{\mu}_\theta(\bm{x}_t,t,\bm{c}) + \sigma_t \bm{z},
\qquad \bm{z}\sim\mathcal{N}(\bm{0},\bm{I}),
\label{eq:ddpm_sampling}
\end{align}
with $\bm{z}=\bm{0}$ typically used at the final step.

\begin{algorithm}[htb!]
\caption{Accelerated Inference with LIM}
\label{alg:fast_infer}
\small

\KwIn{Low-count PET volume $\bm{x}^{\mathrm{LC}}$; \\
pre-trained diffusion model $\epsilon_\theta(\bm{x}_t,t,\bm{x}^{\mathrm{LC}})$ with $L$ blocks $\{m^{(l)}\}_{l=1}^{L}$; \\
Reverse diffusion operator $\mathrm{RevStep}(\cdot)$; \\
Total diffusion steps $T$; starting step $t_s \ll T$; \\
Reuse boundary $K$; \\
Cache interval $N$.}
\KwOut{Reconstructed high-count PET volume $\hat{\bm{x}}_0$.}

\SetKwFunction{RefreshStep}{RefreshStep}
\SetKwFunction{ReuseStep}{ReuseStep}
\SetKwProg{Fn}{Function}{:}{}

\BlankLine
\textbf{Stage 1: Q-Sampling Trajectory Shortcut}\;
Sample $\epsilon \sim \mathcal{N}(0,\mathbf{I})$\;
$\bm{x}_{t_s} \leftarrow \sqrt{\bar{\alpha}_{t_s}}\,\bm{x}^{\mathrm{LC}} + \sqrt{1-\bar{\alpha}_{t_s}}\,\epsilon$\;

\BlankLine
\textbf{Stage 2: Feature-Reuse Network Shortcut}\;
$\mathcal{I} \leftarrow \{\,t_s-iN \mid i \in \mathbb{N},\ 0 \le i \le \lfloor t_s/N \rfloor\,\}$\;
Initialize $\mathbf{F}_{\mathrm{cache}}^{(l)} \leftarrow \varnothing$ for $l=K,\ldots,L$\;

\For{$t \leftarrow t_s$ \KwTo $1$}{
    \uIf{$t \in \mathcal{I}$}{
        $\hat{\epsilon} \leftarrow \RefreshStep(\bm{x}_t,t,\{\mathbf{F}_{\mathrm{cache}}^{(l)}\}_{l=1}^{L})$\;
    }
    \Else{
        $\hat{\epsilon} \leftarrow \ReuseStep(\bm{x}_t,t,\{\mathbf{F}_{\mathrm{cache}}^{(l)}\}_{l=1}^{L},K)$\;
    }
    $\bm{x}_{t-1} \leftarrow \mathrm{RevStep}(\bm{x}_t,t,\hat{\epsilon})$\;
}
$\hat{\bm{x}}_0 \leftarrow \bm{x}_0$\;
\KwRet{$\hat{\bm{x}}_0$}\;

\BlankLine
\Fn{\RefreshStep{$\bm{x}_t,t,\{\mathbf{F}_{\mathrm{cache}}^{(l)}\}_{l=1}^{L}$}}{
    $\mathbf{h}^{(0)} \leftarrow \bm{x}_t$\;
    \For{$l \leftarrow 1$ \KwTo $L$}{
        $\mathbf{h}^{(l)} \leftarrow m^{(l)}(\mathbf{h}^{(l-1)},t)$\;
        $\mathbf{F}_{\mathrm{cache}}^{(l)} \leftarrow \mathbf{h}^{(l)}$\;
    }
    $\hat{\epsilon} \leftarrow \mathrm{Head}(\mathbf{h}^{(L)},t)$\;
    \KwRet{$\hat{\epsilon}$}\;
}

\BlankLine
\Fn{\ReuseStep{$\bm{x}_t,t,\{\mathbf{F}_{\mathrm{cache}}^{(l)}\}_{l=1}^{L},K$}}{
    $\mathbf{h}^{(0)} \leftarrow \bm{x}_t$\;
    \For{$l \leftarrow 1$ \KwTo $K$}{
        $\mathbf{h}^{(l)} \leftarrow m^{(l)}(\mathbf{h}^{(l-1)},t)$\;
    }
    \For{$l \leftarrow K+1$ \KwTo $L$}{
        $\mathbf{h}^{(l)} \leftarrow \mathbf{F}_{\mathrm{cache}}^{(l)}$\;
    }
    $\hat{\epsilon} \leftarrow \mathrm{Head}(\mathbf{h}^{(L)},t)$\;
    \KwRet{$\hat{\epsilon}$}\;
}
\end{algorithm}

\subsection{Redundancy in Diffusion-based PET Denoising}
\label{method:observation}

Although diffusion models have shown strong potential for PET denoising, their iterative reverse process introduces substantial computational overhead. In this work, we observe that diffusion-based PET denoising contains two forms of redundancy: global redundancy along the diffusion trajectory and local redundancy inside the denoising network. These observations motivate our training-free acceleration framework.
\vspace{-0.3cm}
\paragraph{\textbf{Global redundancy along the diffusion trajectory}}
We begin by analyzing the structural relationship between the high-count PET image $\bm{x}^{\mathrm{HC}}$ and its corresponding low-count counterpart $\bm{x}^{\mathrm{LC}}$ under the DDPM forward process.
For a set of five $^{18}$F-FDG PET cases, we apply forward q-sampling
(Eq.~\ref{eq:ddpm_q_xt_x0}) independently to $\bm{x}^{\mathrm{HC}}$ and
$\bm{x}^{\mathrm{LC}}$ at the same step $t$, and measure their cosine similarity across the full diffusion trajectory.

As shown in Fig.~\ref{Fig-Intro}(a), the q-sampled high-count and low-count states remain highly similar throughout the entire forward process, with cosine similarity consistently approaching 1 across all steps.
Notably, even in the original image domain ($t=0$), the cosine similarity between $\bm{x}^{\mathrm{HC}}$ and $\bm{x}^{\mathrm{LC}}$ reaches 0.97, confirming that the low-count PET image already preserves the dominant anatomical structures and coarse tracer distribution of the high-count target.

This observation is consistent with the DDPM forward formulation. Since Eq.~\ref{eq:ddpm_q_xt_x0} constructs $\bm{x}_t$ by perturbing a clean image $\bm{x}_0$ with Gaussian noise, two structurally similar clean images are expected to produce similar noisy samples at the same step. Therefore, the noised low-count PET image can serve as a structurally faithful approximation to the noised high-count PET image at an intermediate step. 
This suggests that the reverse process need not start from pure Gaussian noise: initializing from a q-sampled $\bm{x}^{\mathrm{LC}}$ at an intermediate step $t_s \ll T$ allows the model to bypass the redundant high-noise prefix of the denoising trajectory without discarding meaningful structural
information.

\paragraph{\textbf{Local redundancy inside the denoising network}} We further examine the temporal evolution of intermediate U-Net features during
the reverse process. As shown in Fig.~\ref{Fig-Intro}, features extracted from the encoder, bottleneck, and decoder stages all exhibit high cosine similarity between adjacent reverse denoising steps across the full diffusion trajectory.

This indicates that the denoising network recomputes highly similar intermediate representations at consecutive reverse denoising steps, constituting a substantial
source of redundant computation. High-level decoder features, in particular, evolve smoothly along the reverse trajectory and can therefore be shared across neighboring reverse denoising steps with minimal fidelity degradation. This local redundancy motivates a feature-caching strategy in which selected decoder representations are computed at periodic refresh steps and reused at intervening non-refresh steps, bypassing a portion of the full U-Net forward pass.

\subsection{Q-Sampling Trajectory Shortcut}
\label{method:QTS}

Motivated by the global trajectory redundancy observed in Sec.~\ref{method:observation}, we propose {Q-Sampling Trajectory Shortcut}, a global training-free acceleration strategy that shortens the reverse diffusion trajectory by starting from an intermediate step.

Conditional diffusion models for image-to-image restoration typically initialize sampling from pure Gaussian noise $\bm{x}_T \sim \mathcal{N}(\bm{0}, \bm{I})$ and iteratively denoise the sample over $T$ reverse denoising steps. Although this full reverse trajectory can achieve strong reconstruction quality, it introduces substantial inference cost. In low-count PET denoising, however, the conditional input $\bm{x}^{\mathrm{LC}}$ already preserves the major anatomical structure and coarse tracer distribution of the high-count target. Therefore, the early high-noise reverse denoising steps can be partially bypassed by constructing an intermediate noisy state directly from $\bm{x}^{\mathrm{LC}}$.

Let $\bm{x}^{\mathrm{LC}}$ denote the low-count PET input and $\hat{\bm{x}}_0$ denote the reconstructed high-count PET output. Instead of initializing the reverse process from $\bm{x}_T \sim \mathcal{N}(\bm{0}, \bm{I})$, we construct an intermediate noisy initialization at step $t_s$ using the forward diffusion formulation:
\begin{align}
\bm{x}_{t_s}
=
\sqrt{\bar{\alpha}_{t_s}}\,\bm{x}^{\mathrm{LC}}
+
\sqrt{1-\bar{\alpha}_{t_s}}\,\bm{\epsilon},
\qquad
\bm{\epsilon} \sim \mathcal{N}(\bm{0},\bm{I}).
\label{eq:skip_init}
\end{align}

Starting from $\bm{x}_{t_s}$, we apply the standard conditional reverse update in Eq.~\eqref{eq:ddpm_sampling} for $t=t_s,t_s-1,\ldots,1$ to obtain $\hat{\bm{x}}_0$. This reduces the number of reverse denoising steps from $T$ to $t_s$, directly decreasing inference cost. Since $\bm{x}_{t_s}$ is constructed through the same forward q-sampling formulation used during DDPM training, the initialization remains close to the noisy data distribution encountered by the denoising network. We further discuss why this global skipping strategy may generalize to other diffusion-based image restoration tasks in \ref{Appendix}.

\subsection{Feature-Reuse Network Shortcut}
\label{method:Cache}

Motivated by the local feature redundancy observed in Sec.~\ref{method:observation}, we further accelerate reverse inference by reusing intermediate U-Net features across neighboring reverse denoising steps. Since high-level features evolve smoothly along the reverse trajectory, they can be cached at selected refresh steps and reused during subsequent nearby steps. This allows us to perform partial U-Net inference at non-refresh steps, thereby reducing redundant computation.

\paragraph{Feature reusing with a fixed refresh interval}
Let $E_m(\cdot)$ and $D_m(\cdot)$ denote the $m$-th encoder and decoder blocks of the U-Net, respectively, with a skip connection linking $E_m$ to $D_m$. At a cache-refresh denoising step $t$, we perform a full U-Net forward pass and store the output feature of a selected higher-level decoder block, e.g., $D_{m+1}$:
\begin{align}
\mathbf{F}_{\mathrm{cache}}^{\,t} = \mathbf{h}_{D_{m+1}}^{\,t},
\label{eq:cache_feature}
\end{align}
where $\mathbf{h}_{D_{m+1}}^{\,t}$ denotes the output feature map of decoder block $D_{m+1}$ at denoising step $t$.

For the following reverse denoising steps before the next cache refresh, we reuse the most recently cached feature instead of recomputing the entire U-Net. Specifically, we execute the network only up to the corresponding partial path and construct the input to decoder block $D_m$ by concatenating the newly computed skip feature with the cached decoder feature:
\begin{align}
\mathrm{Input}\!\left(D_m^{t}\right)
=
\mathrm{Concat}\!\left(
\mathbf{h}_{E_m}^{\,t},
\mathbf{F}_{\mathrm{cache}}
\right),
\label{eq:cache_concat}
\end{align}
where $\mathbf{h}_{E_m}^{\,t}$ denotes the skip feature from encoder block $E_m$ at denoising step $t$, and $\mathbf{F}_{\mathrm{cache}}$ denotes the most recently updated cached decoder feature. In this way, the remaining decoder computation above the caching point can be bypassed at non-refresh steps.

To control the efficiency-fidelity trade-off, we adopt a fixed refresh interval $N$. The cache is updated once every $N$ reverse denoising steps using a full U-Net evaluation and reused during the intervening $N-1$ steps with partial inference. Let $T'$ denote the total number of executed reverse denoising steps, where $T'=t_s$ when Q-Sampling Trajectory Shortcut is applied. The cache-refresh step set is defined as
\begin{align}
\mathcal{I}
=
\left\{
t \in \mathbb{N}
\;\middle|\;
t = T' - iN,\;
0 \le i < \left\lceil \frac{T'}{N} \right\rceil
\right\}.
\label{eq:cache_schedule}
\end{align}

For each $t \in \mathcal{I}$, a full U-Net forward pass is performed and the cache is updated according to Eq.~\eqref{eq:cache_feature}. For each $t \notin \mathcal{I}$, the most recently cached feature is reused, and only partial U-Net inference is executed according to Eq.~\eqref{eq:cache_concat}. The hyperparameters $(m, N)$ explicitly control the trade-off between inference efficiency and denoising fidelity.

\begin{figure*}[htb!]
    \centering
    \includegraphics[width=0.95\textwidth]{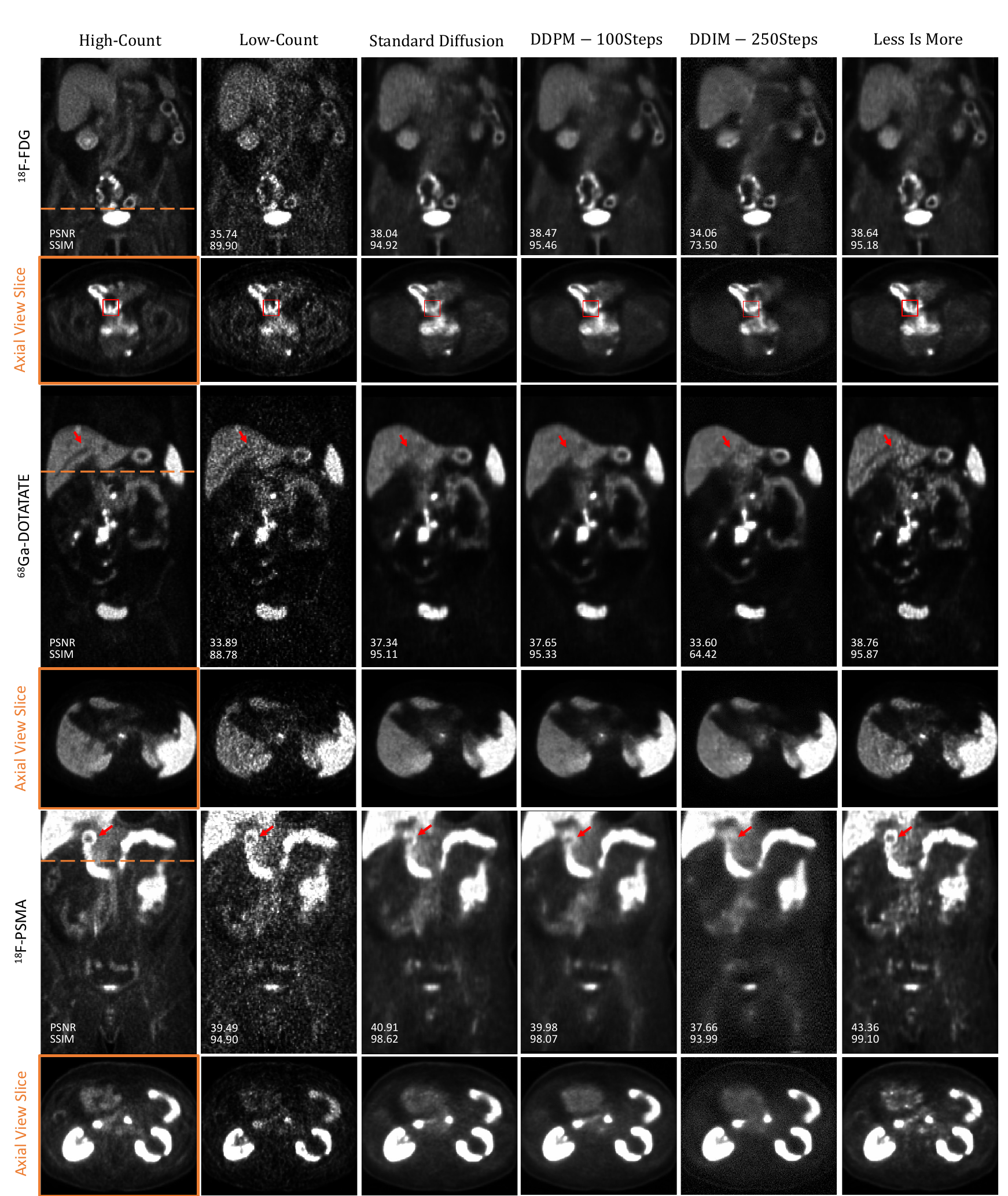}
    \vspace{-8pt}
    \caption{Visualization on the three in-house PET tracer cohorts acquired at Northwestern Memorial Hospital, including $^{18}$F-FDG, $^{68}$Ga-DOTATATE, and $^{18}$F-PSMA PET. Representative coronal and axial views are shown for the high-count reference, low-count input, Standard Diffusion, accelerated scheduler baselines, and the proposed LIM that was applied on top of Standard Diffusion. Red arrows indicate representative regions with visible differences in structural preservation and tracer uptake patterns. Across all tracers, LIM preserves sharper anatomical boundaries and lesion-related uptake while maintaining effective noise suppression. Notably, although the pre-trained diffusion model was originally developed using public $^{18}$F-FDG data, the proposed acceleration strategy remains effective on unseen tracer distributions ($^{68}$Ga-DOTATATE and $^{18}$F-PSMA), suggesting favorable generalization without retraining.}
    \label{Fig-Viz-Three-Tracer-Internal}
\end{figure*}

\begin{table*}[htb!]
\centering
\caption{Quantitative comparison of diffusion acceleration strategies on three in-house PET tracer cohorts from Northwestern Memorial Hospital, including $^{18}$F-FDG ($n=15$), $^{68}$Ga-DOTATATE ($n=32$), and $^{18}$F-PSMA ($n=25$). All methods were evaluated using the same pre-trained 3D diffusion denoising model without retraining or fine-tuning. Results are reported as mean $\pm$ standard deviation. Best performance within each metric and tracer cohort is highlighted in \textbf{bold}.}
\label{tab:three_tracers_baseline_scheduler_cache}

\renewcommand{\arraystretch}{1.10}
\setlength{\tabcolsep}{3.2pt}
\scriptsize

\begin{tabularx}{\textwidth}{@{}l *{9}{Y}@{}}
\toprule
& \multicolumn{3}{c}{\textbf{$^{18}$F-FDG} 15 Cases}
& \multicolumn{3}{c}{\textbf{$^{68}$Ga-DOTATATE} 32 Cases}
& \multicolumn{3}{c}{\textbf{$^{18}$F-PSMA} 25 Cases} \\
\cmidrule(lr){2-4}\cmidrule(lr){5-7}\cmidrule(lr){8-10}
\textbf{Method / Setting}
& \textbf{PSNR}$\uparrow$ & \textbf{SSIM}($\times10^2$)$\uparrow$ & \textbf{MAE}($\times10^{-2}$)$\downarrow$
& \textbf{PSNR}$\uparrow$ & \textbf{SSIM}($\times10^2$)$\uparrow$ & \textbf{MAE}($\times10^{-2}$)$\downarrow$
& \textbf{PSNR}$\uparrow$ & \textbf{SSIM}($\times10^2$)$\uparrow$ & \textbf{MAE}($\times10^{-2}$)$\downarrow$ \\
\midrule

\multicolumn{10}{@{}l@{}}{\textbf{Input}}\\[-2pt]
10\% low-count
& 38.88$\pm$4.24 & 93.05$\pm$3.38 & 7.342$\pm$1.583
& 39.19$\pm$4.05 & 93.63$\pm$3.42 & 8.508$\pm$3.078
& 40.89$\pm$2.99 & 95.40$\pm$2.23 & 9.990$\pm$2.960 \\

\midrule

\multicolumn{10}{@{}l@{}}{\textbf{Baseline \& Quantization}}\\[-2pt]
Standard Diffusion
& 41.58$\pm$1.92 & 97.24$\pm$0.70 & 5.269$\pm$1.109
& 42.21$\pm$3.54 & 97.18$\pm$1.40 & 5.390$\pm$1.700
& 42.05$\pm$2.42 & 97.97$\pm$0.59 & 6.000$\pm$1.637 \\
Standard Diffusion (BF16)
& 41.50$\pm$2.02 & 97.23$\pm$0.70 & 5.273$\pm$1.114
& 42.19$\pm$3.52 & 97.17$\pm$1.41 & 5.417$\pm$1.674
& 42.04$\pm$2.43 & 97.95$\pm$0.83 & 7.081$\pm$1.684 \\
\midrule

\multicolumn{10}{@{}l@{}}{\textbf{Scheduler Replacement}}\\[-2pt]
DDPM-100 Steps
& 41.29$\pm$4.09 & 96.85$\pm$1.62 & \textbf{4.982$\pm$1.094}
& 42.50$\pm$3.57 & 97.37$\pm$1.32 & 5.249$\pm$1.648
& 43.58$\pm$2.72 & 98.29$\pm$0.59 & 6.841$\pm$1.662 \\
DDIM-250 Steps
& 37.40$\pm$3.10 & 81.46$\pm$9.53 & 19.168$\pm$13.935
& 34.82$\pm$1.26 & 62.26$\pm$5.87 & 24.549$\pm$10.722
& 38.01$\pm$1.35 & 86.12$\pm$2.84 & 18.896$\pm$4.694 \\
\midrule

\multicolumn{10}{@{}l@{}}{\textbf{Ours}}\\[-2pt]
Less Is More-Q
& 42.10$\pm$1.87\textsuperscript{$\dagger$} & 97.33$\pm$0.68\textsuperscript{$\diamond$} & 5.110$\pm$1.866\textsuperscript{$\ddagger$}
& $43.78\pm3.71^{\dagger}$ 
& $97.62\pm1.20^{\dagger}$ 
& $4.752\pm1.576^{\dagger}$ 
& $44.68\pm$2.84$^{\dagger}$ 
& $98.37\pm$0.76$^{\dagger}$ 
& $5.793\pm$1.501$^{\dagger}$ \\
Less Is More-Full
& \textbf{42.18$\pm$1.84}\textsuperscript{$\dagger$} & \textbf{97.34$\pm$0.68}\textsuperscript{$\diamond$} & 5.147$\pm$1.099\textsuperscript{*}
& $\mathbf{43.85\pm3.71}^{\dagger}$ 
& $\mathbf{97.65\pm1.18}^{\dagger}$ 
& $\mathbf{4.752\pm1.576}^{\dagger}$ 
& $\mathbf{44.78\pm2.82}^{\dagger}$ 
& $\mathbf{98.40\pm0.74}^{\dagger}$ 
& $\mathbf{5.774\pm1.473}^{\dagger}$ \\
\bottomrule
\end{tabularx}

\vspace{2pt}
\scriptsize
\noindent\textit{Notes.}
Less Is More-Q denotes the tracer-specific optimal configuration using only the Q-Sampling Trajectory Shortcut, without feature reuse. The starting denoising step $t_s$ was set to 50 for $^{18}$F-FDG, 25 for $^{68}$Ga-DOTATATE, and 10 for $^{18}$F-PSMA. Less Is More-Full denotes the full configuration that combines the Q-Sampling Trajectory Shortcut with the Feature-Reuse Network Shortcut, using a feature-reuse interval of $N=3$. Statistical significance was assessed against Standard Diffusion using a two-sided paired Wilcoxon signed-rank test: * $p<0.05$, $\diamond$ $p<0.01$, $\ddagger$ $p<0.001$, $\dagger$ $p<0.0001$.
\end{table*}

\subsection{Dataset Description}
We evaluated the proposed framework on both in-house and public PET datasets encompassing multiple tracers and acquisition protocols. The in-house dataset, collected at Northwestern Memorial Hospital, includes three clinically important tracers: $^{18}$F-FDG, $^{68}$Ga-DOTATATE, and $^{18}$F-PSMA. To further assess generalizability, we additionally evaluated the proposed method on a publicly available whole-body $^{18}$F-FDG PET dataset from the UDPET Challenge~\citep{UDPET_Dataset,cross-tracerPETDataset}. Together, these datasets enable comprehensive evaluation across different tracer distributions, patient populations, and imaging systems.

\paragraph{\textbf{In-house dataset}} We retrospectively collected three clinical PET cohorts from Northwestern Memorial Hospital, including 15 $^{18}$F-FDG studies, 32 $^{68}$Ga-DOTATATE studies, and 25 $^{18}$F-PSMA studies. For each tracer, low-count PET images were generated by uniformly downsampling the original list-mode data with a dose reduction factor (DRF) of 10. Both low-count and corresponding high-count PET images were reconstructed using the same ordered-subsets expectation maximization (OSEM) protocol, with consistent attenuation, scatter, and random-event corrections applied across all reconstructions. The resulting paired low-count and high-count PET images served as the input and reference data, respectively, for quantitative evaluation. Reconstructed images had a matrix size of $440 \times 440$ in the transverse plane, while the axial dimension varied according to tracer-specific acquisition protocols and patient characteristics.

\paragraph{\textbf{Public dataset}} To evaluate robustness on an external dataset, we further conducted experiments on the UDPET Challenge dataset~\citep{UDPET_Dataset}, a publicly available benchmark for ultra-low-count whole-body PET restoration. We randomly selected 38 $^{18}$F-FDG PET studies acquired on the United Imaging uExplorer total-body PET/CT scanner. PET images were reconstructed using OSEM with three-dimensional time-of-flight (TOF) and point-spread-function (PSF) modeling. The dataset provides paired low-count and high-count PET images generated through retrospective list-mode subsampling at multiple dose reduction factors (DRFs = 4, 10, 20, 50, and 100). Because each low-count image is reconstructed from a subset of the same acquisition used to generate the corresponding high-count reference, the image pairs are intrinsically aligned and well suited for supervised PET denoising evaluation. The reconstructed voxel size was $1.667 \times 1.667 \times 2.886~\mathrm{mm}^{3}$.

\subsection{Implementation Details}
Our pipeline was implemented in PyTorch, and all experiments were conducted on a single NVIDIA A100 80GB GPU. Unless otherwise specified, inference was performed with a batch size of 1 using FP16 precision. We directly used the model with the pre-trained weights released by ~\cite{yu2025robust}, which were originally developed for robust whole-body PET image denoising using 3D diffusion models. For each PET volume $\bm{x}$, we applied a per-volume intensity normalization scheme. Specifically, we first computed a robust maximum $m = \mathrm{P}_{99.99}(\bm{x})$, where $\mathrm{P}_{99.99}(\cdot)$ denotes the 99.99th percentile of voxel intensities. We then clipped the volume to suppress extreme outliers and normalized it to $[0,1]$. During inference, the network predicts a normalized output $\hat{\bm{x}}^{\mathrm{norm}}$, which is mapped back to the original intensity scale by
$\hat{\bm{x}} = m \cdot \hat{\bm{x}}^{\mathrm{norm}}$. The same preprocessing and inverse rescaling procedure was used for all tracers. Whole-volume inference was performed using a sliding-window strategy with overlap. Each cropped volume was decomposed into fixed-size 3D patches of size $224 \times 224 \times 96$, which is identical to the training patch size. Each patch was processed independently by the network, and the patch-wise predictions were stitched back to the original cropped volume using linear blending in the overlapping regions.

\subsection{Baselines and Evaluation Strategies}
\label{sec:Baselines_And_Eval_Strategies}

To ensure a controlled and fair comparison, all acceleration methods were evaluated using the same pre-trained 3D diffusion denoising model released by~\cite{yu2025robust}, with identical low-count PET inputs and corresponding high-count references. No additional training, fine-tuning, or architectural modification was performed for any compared method. This experimental design isolates the effect of inference-time acceleration strategies while avoiding confounding factors introduced by differences in model capacity, training protocols, or optimization settings.

We intentionally focus on \emph{training-free acceleration} rather than comparing against newly trained restoration models such as 3D U-Net or GAN variants. Training large-scale 3D restoration networks for whole-body PET imaging is computationally intensive due to the high spatial resolution and volumetric memory requirements. For example,~\cite{yu2025robust} reported that training the reference 3D DDPM required 8 NVIDIA A100 GPUs over 21 days using 302 whole-body PET cases. Moreover, the original work already demonstrated superior restoration performance relative to representative restoration baselines, including 2.5D DDPM, 3D U-Net, and 3D GAN models. Therefore, instead of retraining additional restoration methods, this study investigates a more targeted and practically relevant question: \emph{whether an already high-performing 3D diffusion model can be substantially accelerated at inference time while preserving reconstruction fidelity.} Recent feature-reuse acceleration methods have predominantly been developed for text-to-image~\citep{labs2025flux1kontextflowmatching} and text-to-video~\citep{WAN} generation, where diffusion backbones are commonly implemented using Diffusion Transformers (DiTs). These approaches often exploit transformer-specific properties such as token-level redundancy~\citep{ToCa}, attention-map stability~\citep{Jenga}, and timestep-dependent activation reuse~\citep{chung2026seacachespectralevolutionawarecacheaccelerating}. However, such assumptions are not directly applicable to our setting, where the denoising backbone is a volumetric 3D U-Net composed of hierarchical feature maps and encoder--decoder skip connections. Consequently, feature reuse for whole-body PET restoration requires a U-Net-specific design that accounts for volumetric representations and memory constraints.

Based on this setup, we evaluated the following inference strategies: \textbf{(1) Standard Diffusion.} The original inference procedure of the pre-trained 3D diffusion model using the complete reverse trajectory ($T=1000$) and standard precision. This serves as the primary reference for both image quality and computational cost. \textbf{(2) Standard Diffusion (BF16).} The same 1000-step reverse process executed using bfloat16 mixed precision to evaluate the benefit of numerical precision reduction alone. \textbf{(3) DDPM-100 Steps.} A reduced-step baseline that uniformly shortens the reverse trajectory to 100 stochastic DDPM reverse denoising steps without modifying the pre-trained model. \textbf{(4) DDIM-250 Steps.} A deterministic sampling baseline using DDIM with 250 reverse denoising steps, representing a commonly adopted training-free diffusion acceleration strategy. 

To evaluate the trade-off between reconstruction fidelity and computational efficiency, we report both image-quality and efficiency metrics. Image quality was assessed using peak signal-to-noise ratio (PSNR), structural similarity index (SSIM), and mean absolute error (MAE). Computational performance was evaluated using inference time and computational complexity.

\paragraph{\textbf{Blinded Reader Study}} To further assess clinical utility beyond quantitative image metrics, we conducted a blinded reader study using the three in-house tracer cohorts: $^{18}$F-FDG, $^{68}$Ga-DOTATATE, and $^{18}$F-PSMA. For each tracer, 10 PET studies were randomly selected, resulting in 30 evaluated cases. Each case was reviewed together with the corresponding CT image and four PET image sets: (i) high-count PET, (ii) low-count PET, (iii) reconstruction generated by Standard Diffusion, and (iv) reconstruction generated by the proposed LIM framework. High-count and low-count PET images were explicitly labeled to provide clinical context, whereas the two reconstructed images were anonymized and randomly ordered such that readers remained blinded to the reconstruction method.

Readers independently evaluated reconstructed PET images using a 5-point Likert scale across three clinically relevant criteria: noise suppression, clinical confidence, and lesion visibility, where higher scores indicate better perceived image quality and diagnostic utility. In addition, readers provided an overall pairwise preference between the two reconstruction methods using four categories: \emph{Baseline is better}, \emph{Baseline is slightly better}, \emph{Ours is slightly better}, and \emph{Ours is better}. Cases judged to be visually indistinguishable were recorded as ties and excluded from preference ranking analysis.

\begin{figure*}[htb!]
    \centering
    \includegraphics[width=0.94\textwidth]{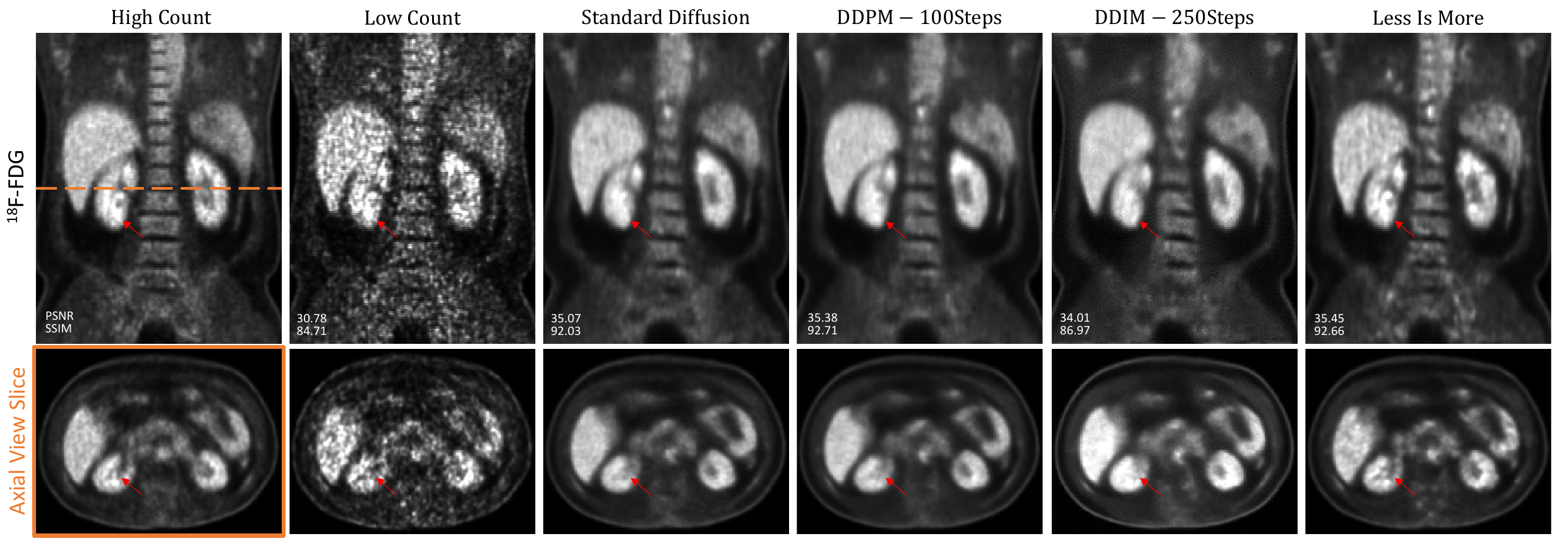}
    \vspace{-8pt}
    \caption{Visualization on the public UDPET $^{18}$F-FDG PET dataset. The proposed Less Is More pipeline better restores lesion details in the kidney region while maintaining substantially shorter inference time. All methods achieve visually plausible denoising results in this case, suggesting that the public data are reasonably aligned with the training distribution. Nevertheless, Less Is More achieves the most favorable efficiency while preserving competitive structural fidelity.}
    \label{fig-external-dataset}
\end{figure*}

\begin{table*}[htb!]
\centering
\caption{Quantitative comparison and computational analysis on the public UDPET $^{18}$F-FDG PET dataset. All methods were evaluated using the same pre-trained 3D diffusion denoising model without retraining or fine-tuning. Results are reported as mean $\pm$ standard deviation. Best performance within each metric is highlighted in \textbf{bold}.}
\label{tab:computational_complexity}

\renewcommand{\arraystretch}{1.12}
\setlength{\tabcolsep}{5pt}
\scriptsize

\begin{tabular}{@{}lccccc@{}}
\toprule
\textbf{Method / Setting} &
\textbf{PSNR}$\uparrow$ &
\textbf{SSIM}($\times 10^2$)$\uparrow$ &
\textbf{MAE}($\times 10^{-2}$)$\downarrow$ &
\textbf{TFLOPs}$\downarrow$ &
\textbf{Inference Time}$\downarrow$ \\
\midrule

\multicolumn{6}{@{}l@{}}{\textbf{Input}}\\[-2pt]
10\% low-count
& 37.85$\pm$4.71
& 93.43$\pm$3.49
& 4.611$\pm$1.340
& --
& -- \\
\midrule

\multicolumn{6}{@{}l@{}}{\textbf{Baseline \& Quantization}}\\[-2pt]
Standard Diffusion
& 40.17$\pm$3.61
& 96.44$\pm$2.00
& 3.240$\pm$0.865
& 107149.7
& 56.62 min \\
Standard Diffusion (BF16)
& 40.19$\pm$3.63
& 96.44$\pm$2.00
& 3.253$\pm$0.857
& 107149.7
& 46.23 min \\
\midrule

\multicolumn{6}{@{}l@{}}{\textbf{Schedulers Replacement}}\\[-2pt]
DDPM-100 Steps
& 40.56$\pm$3.68
& 96.81$\pm$1.83
& 3.066$\pm$0.826
& 10714.9
& 5.66 min \\
DDIM-250 Steps
& 36.74$\pm$2.22
& 86.14$\pm$2.88
& 7.232$\pm$3.664
& 26787.3
& 14.16 min \\
\midrule

\multicolumn{6}{@{}l@{}}{\textbf{Ours}}\\[-2pt]
Less Is More-Q
& $\mathbf{41.05\pm3.93}^{\dagger}$
& $\mathbf{96.78\pm1.79}^{\dagger}$
& $\mathbf{3.044\pm0.839}^{\dagger}$
& 5357.5
& 2.83 min \\
Less Is More-Full
& $41.03\pm3.79^{\dagger}$
& $96.48\pm1.73^{ns}$
& $3.428\pm0.958^{ns}$
& \textbf{2539.6}
& \textbf{1.48 min} \\
\bottomrule
\end{tabular}

\vspace{2pt}
\footnotesize
\noindent\textit{Notes.} Computational complexity is measured over the entire denoising process for a single PET volume. Since all methods are built upon the same pre-trained model, they share the same number of model parameters. \textbf{Less Is More-Q} denotes the optimal early-start configuration without feature reuse, with the starting step set to $t_s=50$ for $^{18}$F-FDG. \textbf{Less Is More-Full} denotes the combination of the optimal early-start configuration and feature reuse with interval $N=3$. Statistical significance was assessed against Standard Diffusion using a two-sided paired Wilcoxon signed-rank test: ns $p>0.05$, * $p<0.05$, $\diamond$ $p<0.01$, $\ddagger$ $p<0.001$, $\dagger$ $p<0.0001$.
\label{tab-external-dataset}
\end{table*}


\section{Experiments and Results}
\label{E&R}

\subsection{Qualitative Results}
\label{Visualization Results}

We first present qualitative comparisons to assess whether the proposed acceleration strategy preserves clinically meaningful image characteristics while substantially shortening inference. Fig.~\ref{Fig-Viz-Three-Tracer-Internal} shows representative reconstruction results on the three in-house tracer cohorts, including $^{18}$F-FDG, $^{68}$Ga-DOTATATE, and $^{18}$F-PSMA PET. For each tracer, we visualize both coronal and axial views and compare the high-count reference, low-count input, reconstruction generated by the original pre-trained diffusion model~\cite{yu2025robust}, and the proposed LIM framework using the optimal configuration ($t_s$ and $N=3$).

Overall, both Standard Diffusion and LIM substantially improve image quality relative to the low-count input. However, visual inspection reveals notable differences in reconstruction characteristics. Standard Diffusion tends to generate smoother uptake distributions, occasionally suppressing fine lesion boundaries and local uptake variations. In contrast, LIM preserves sharper structural details and more heterogeneous tracer uptake patterns while maintaining effective denoising. This behavior becomes particularly evident for $^{68}$Ga-DOTATATE and $^{18}$F-PSMA PET, where the pre-trained model was originally optimized using public $^{18}$F-FDG data. Despite this cross-tracer distribution shift, LIM maintains favorable restoration quality and preserves clinically relevant uptake structures, suggesting improved robustness under unseen tracer distributions.

We further evaluated the proposed framework on an external public $^{18}$F-FDG PET dataset. Representative examples are shown in Fig.~\ref{fig-external-dataset}. Since the public dataset more closely matches the training distribution of the pre-trained model, all methods produce visually plausible denoising results. Nevertheless, LIM better preserves local lesion structures and uptake heterogeneity, particularly in the kidney region, while substantially reducing inference time.

\begin{figure*}[htb!]
    \centering
    \includegraphics[width=0.93\textwidth]{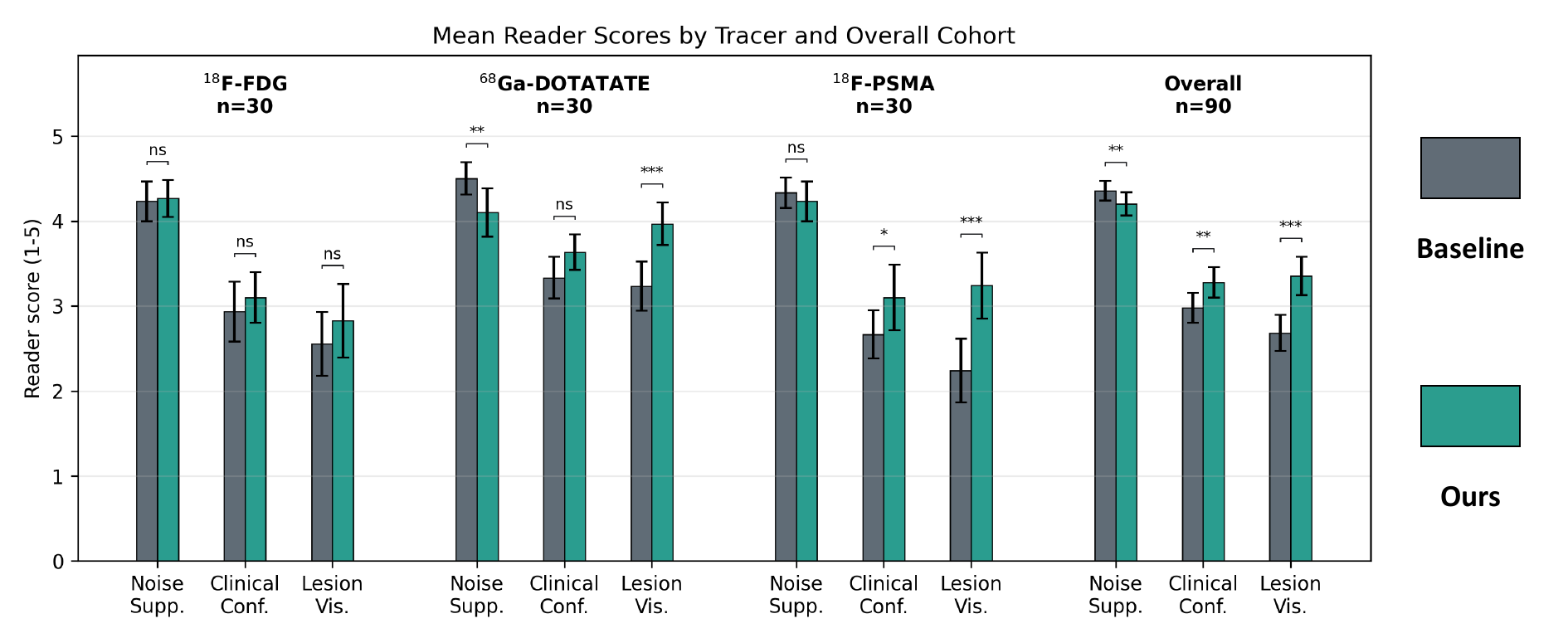}
    \vspace{-8pt}
    \caption{Results of the blinded reader study comparing \textbf{Standard Diffusion \citep{yu2024pet}} (Baseline) and the proposed \textbf{Less Is More (LIM)} framework across tracer-specific and overall cohorts. Mean reader scores are shown for \emph{noise suppression}, \emph{clinical confidence}, and \emph{lesion visibility}. Evaluation was performed separately on $^{18}$F-FDG, $^{68}$Ga-DOTATATE, and $^{18}$F-PSMA PET and additionally aggregated across all cohorts (right). Error bars denote 95\% confidence intervals. LIM achieved comparable noise suppression while improving perceived clinical confidence and lesion visibility across multiple cohorts. Statistical significance was assessed using paired two-sided t-tests. ns: $p \geq 0.05$; *: $p < 0.05$; **: $p < 0.01$; ***: $p < 0.001$.}
    \label{fig:reader-study-mean-score}
\end{figure*}

\begin{figure}[t]
    \centering
    \includegraphics[width=0.93\columnwidth]{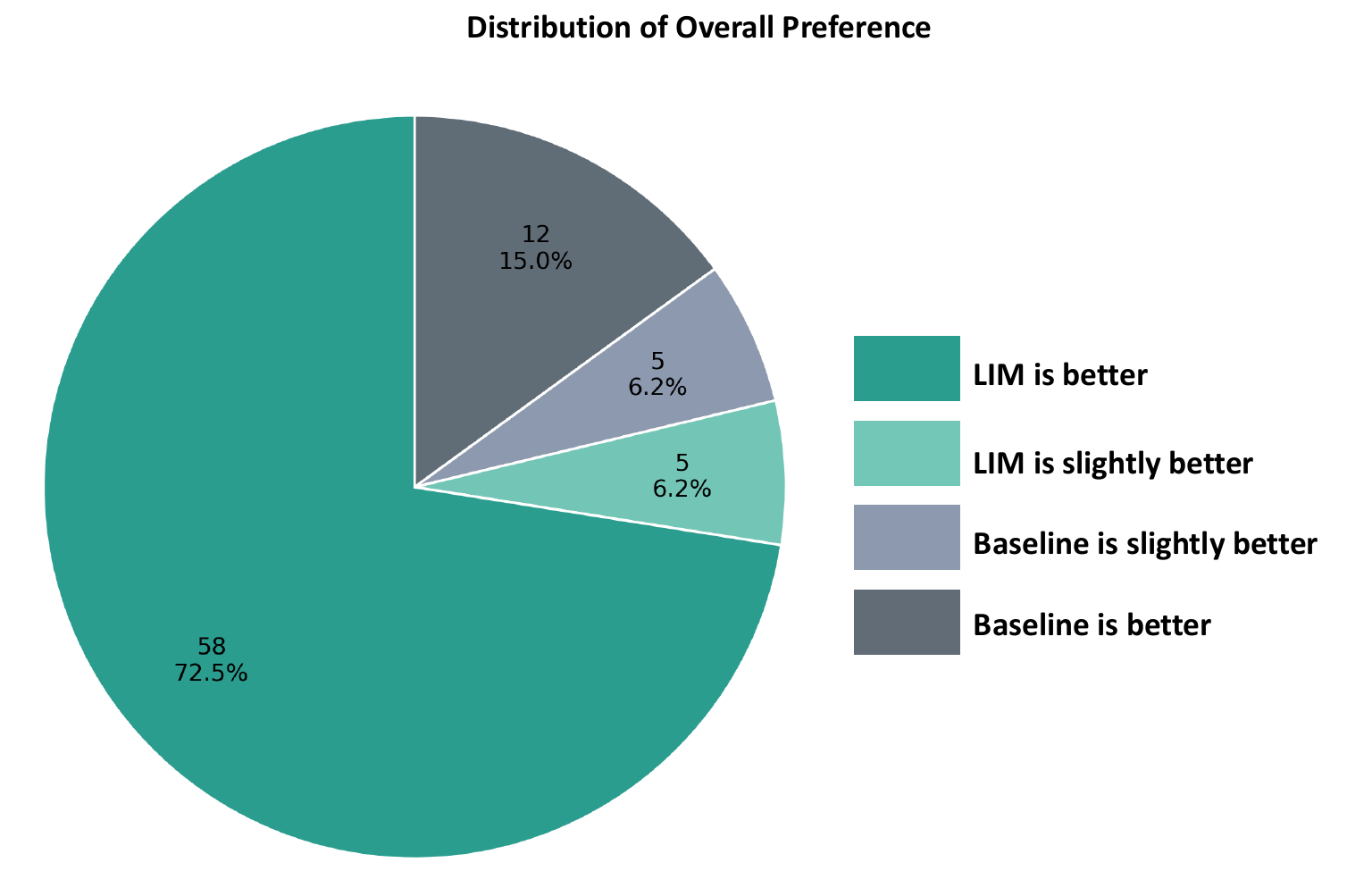}
    \vspace{-8pt}
    \caption{Overall preference distribution from the blinded reader study comparing \textbf{Standard Diffusion} (Baseline) and the proposed \textbf{Less Is More (LIM)} framework. Each sector summarizes the proportion of reader evaluations favoring either method across four ordered categories: \emph{Baseline is better}, \emph{Baseline is slightly better}, \emph{LIM is slightly better}, and \emph{LIM is better}. Values inside the pie chart indicate counts and percentages of all non-tie evaluations, illustrating that LIM was preferred in the majority of comparisons.}
    \label{fig:reader-study-overall-ranking}
\end{figure}

\subsection{Quantitative Results}
\label{E&R: Quantitative Evaluation}

Table~\ref{tab:three_tracers_baseline_scheduler_cache} summarizes quantitative results across the three in-house PET datasets. Overall, the original diffusion model already achieves strong restoration performance across all tracers, confirming the effectiveness of diffusion-based PET denoising. Reducing numerical precision alone through BF16 produces negligible changes in reconstruction quality, indicating that precision reduction mainly affects hardware utilization rather than denoising behavior.

Replacing the original sampling procedure with shorter diffusion schedules yields mixed results. DDPM-100 Steps achieves substantial acceleration but does not consistently improve image fidelity. More aggressive scheduler replacement using DDIM-250 Steps leads to pronounced degradation across all metrics, particularly for $^{68}$Ga-DOTATATE PET, where PSNR decreases from 42.21 to 34.82 and SSIM drops from 97.18 to 62.26. These findings suggest that naively shortening the reverse trajectory may alter the behavior of the pre-trained diffusion process and compromise restoration quality. In contrast, the proposed LIM framework consistently achieves the strongest quantitative performance across all tracer cohorts. LIM-Full achieves the highest PSNR and SSIM on all three datasets while maintaining competitive MAE. Compared with Standard Diffusion, the improvements are statistically significant for multiple metrics. Interestingly, acceleration not only preserves performance but occasionally improves reconstruction quality, suggesting that removing redundant portions of the diffusion trajectory may reduce accumulated denoising artifacts.

Consistent findings are observed on the external public dataset (Table~\ref{tab-external-dataset}). LIM-Q achieves the strongest reconstruction fidelity overall, improving PSNR from 40.17 dB to 41.05 dB while simultaneously reducing inference latency. LIM-Full further improves computational efficiency with only minor changes in quantitative performance.

\begin{table}[htb!]
\centering
\caption{Inference efficiency comparison across three in-house PET tracer cohorts from Northwestern Memorial Hospital, including $^{18}$F-FDG, $^{68}$Ga-DOTATATE, and $^{18}$F-PSMA. All methods were evaluated using the same pre-trained 3D diffusion denoising model without retraining or fine-tuning. Best efficiency within each tracer cohort is highlighted in \textbf{bold}.}
\label{tab:inference_efficiency_three_tracers}

\renewcommand{\arraystretch}{1.08}
\setlength{\tabcolsep}{3.2pt}
\scriptsize

\begin{tabular}{@{}lccc@{}}
\toprule
\textbf{Method}
& \textbf{$^{18}$F-FDG}
& \textbf{$^{68}$Ga-DOTATATE}
& \textbf{$^{18}$F-PSMA} \\
\midrule
Standard Diffusion
& 56.6 min / 1.0$\times$
& 84.9 min / 1.0$\times$
& 56.6 min / 1.0$\times$ \\

Standard Diffusion (BF16)
& 46.2 min / 1.2$\times$
& 69.3 min / 1.2$\times$
& 46.2 min / 1.2$\times$ \\

DDPM-100  Steps
& 5.7 min / 10.0$\times$
& 8.6 min / 10.0$\times$
& 5.7 min / 10.0$\times$ \\

DDIM-250 Steps
& 14.1 min / 4.0$\times$
& 21.2 min / 4.0$\times$
& 14.1 min / 4.0$\times$ \\

Less Is More-Q
& 2.9 min / 20.0$\times$
& 2.1 min / 40.0$\times$
& 0.6 min / 100.0$\times$ \\

Less Is More-Full
& \textbf{1.3 min / 43.5$\times$}
& \textbf{1.0 min / 84.9$\times$}
& \textbf{0.3 min / 188.7$\times$} \\
\bottomrule
\label{tab:in-house_data_efficiency}
\end{tabular}

\vspace{2pt}
\scriptsize
\noindent\textit{Notes.}
Each entry reports inference time in minutes per 3D PET volume and speedup relative to Standard Diffusion within the same tracer.
LIM-Q denotes Less Is More with the tracer-specific Q-Sampling Trajectory Shortcut only.
LIM-Full further combines the Feature-Reuse Network Shortcut with $N=3$.
\end{table}

\subsection{Computational Efficiency and Complexity}
\label{Sec-Memory_And_Computational_Complexity}

We next evaluate whether acceleration translates into practical efficiency gains. Inference speed on the in-house datasets is summarized in Table~\ref{tab:inference_efficiency_three_tracers}, while computational complexity and timing on the public dataset are reported in Table~\ref{tab-external-dataset}. Standard Diffusion requires approximately 57–85 minutes per whole-body PET volume, corresponding to over $10^5$ TFLOPs per inference. Although BF16 reduces latency by approximately 18\%, the computational burden remains dominated by repeated diffusion iterations.

Scheduler replacement reduces runtime but introduces substantial quality degradation. In comparison, LIM provides significantly better quality–efficiency tradeoffs. LIM-Q reduces inference time to 2.83 minutes while improving quantitative fidelity, corresponding to approximately $20\times$ acceleration. Further incorporating feature reuse yields LIM-Full, reducing computational complexity from 107149.7 TFLOPs to 2539.6 TFLOPs and shortening inference to 1.48 minutes.

Across tracer cohorts, LIM-Full achieves acceleration factors of up to $43.5\times$, $84.9\times$, and $188.7\times$, respectively. These results demonstrate that trajectory reduction and feature reuse operate synergistically to reduce both the total number of denoising steps and the computational burden of each remaining step.

\begin{figure}[htb!]
    \centering
    \includegraphics[width=\columnwidth]{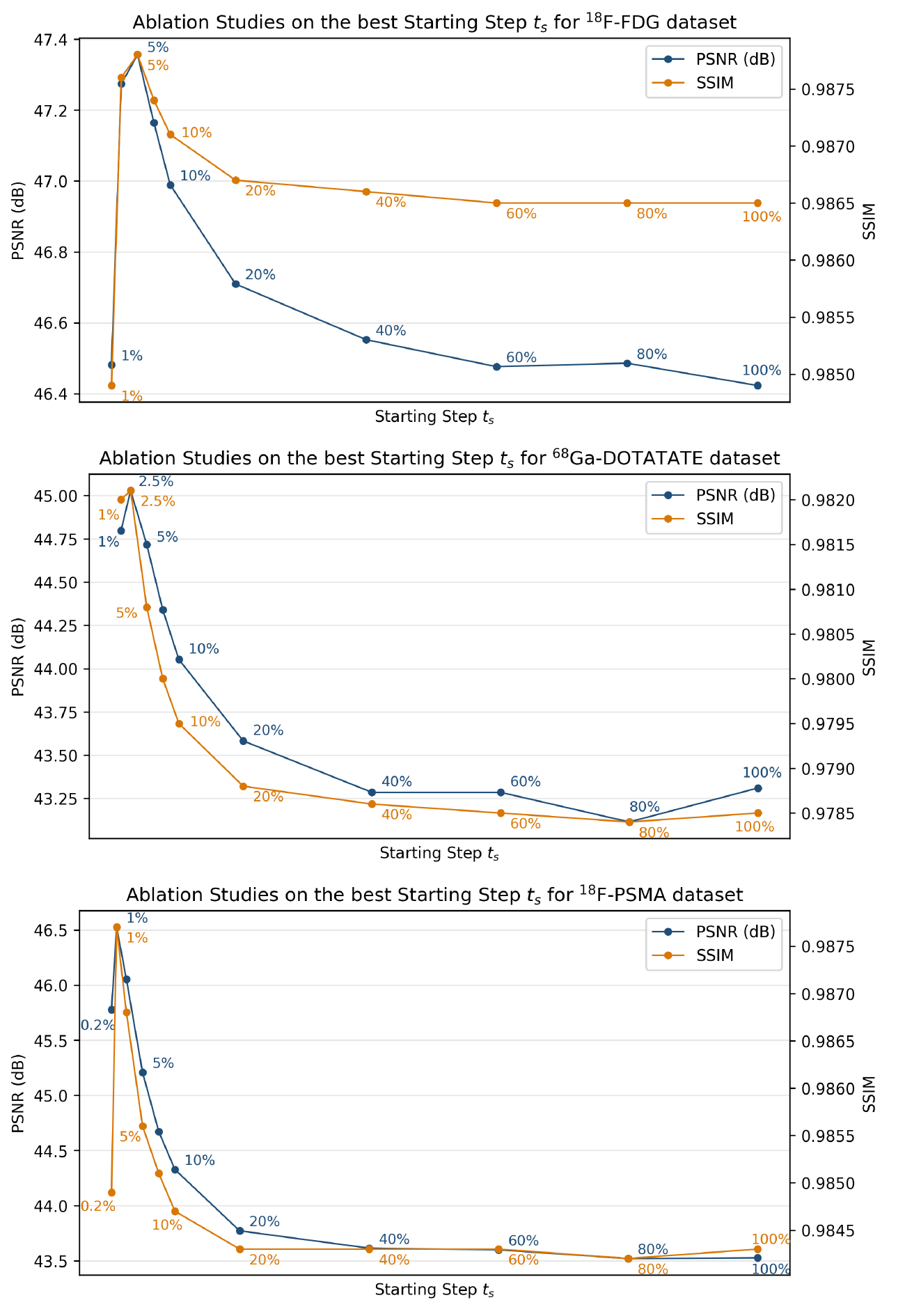}
    \vspace{-8pt}
    \caption{Ablation study of the starting denoising timestep $t_s$ for the proposed Q-Sampling Trajectory Shortcut on $^{18}$F-FDG, $^{68}$Ga-DOTATATE, and $^{18}$F-PSMA PET. PSNR and SSIM are evaluated across different trajectory initialization ratios (expressed as a percentage of the full reverse diffusion trajectory). Performance initially improves as redundant early denoising steps are removed, reaches an optimum at intermediate starting points, and gradually degrades when the trajectory is shortened excessively. The optimal configurations occur at 5\% ($t_s=50$) for $^{18}$F-FDG, 2.5\% ($t_s=25$) for $^{68}$Ga-DOTATATE, and 1\% ($t_s=10$) for $^{18}$F-PSMA, suggesting tracer-dependent redundancy in the reverse diffusion process.}
    \label{fig:ablation-t_s}
\end{figure}

\begin{figure}[htb!]
\label{fig:Ablation-Speed}
    \centering
    \includegraphics[width=0.92\columnwidth]{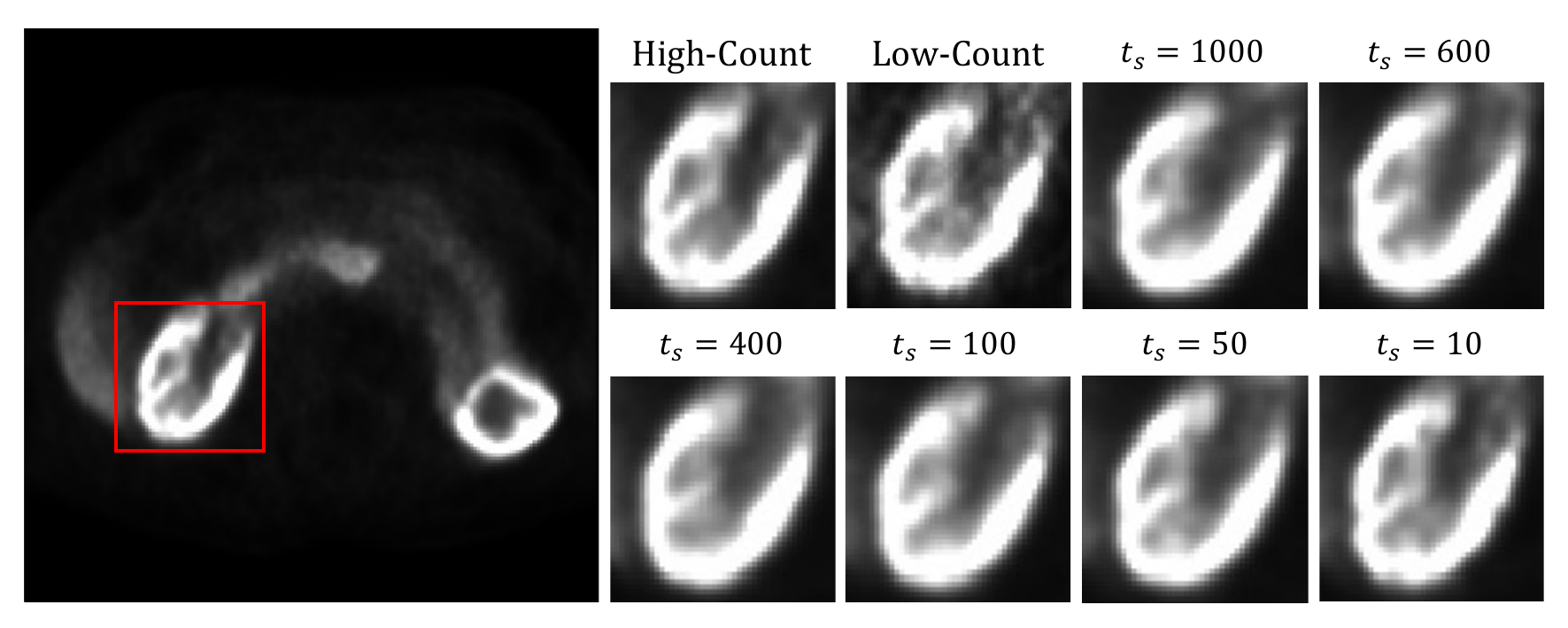}
    \vspace{-8pt}
    \caption{Visualization results under different starting denoising timesteps $t_s$ for 10\% low-count $^{18}$F-PSMA PET denoising. The red box indicates the zoomed region for comparison. Starting reverse diffusion from an intermediate timestep preserves clearer structural details and lesion morphology while reducing redundant denoising iterations. Excessive trajectory shortening, however, leads to degradation in anatomical fidelity.}
    \label{Fig-Ablation-t_s}
\end{figure}

\subsection{Blinded Reader Study}
\label{E&R: Blind reader study}

To assess clinical utility beyond image-level metrics, we conducted a blinded reader study following the protocol described in Sec.~\ref{sec:Baselines_And_Eval_Strategies}. Fig.~\ref{fig:reader-study-mean-score} summarizes reader ratings across tracer-specific and pooled cohorts. Overall, LIM received significantly higher scores for clinical confidence and lesion visibility compared with Standard Diffusion, whereas Standard Diffusion achieved slightly higher noise suppression ratings.

These findings align with the qualitative observations in Fig.~\ref{Fig-Viz-Three-Tracer-Internal} and Fig.~\ref{fig-external-dataset}. Standard Diffusion frequently produced smoother uptake distributions and reduced visible image noise but occasionally suppressed lesion boundaries and local uptake variation. In contrast, LIM retained moderate physiological texture while preserving lesion conspicuity and anatomical detail, resulting in improved perceived diagnostic confidence. Fig.~\ref{fig:reader-study-overall-ranking} summarizes overall reader preference. Across 90 reader-case evaluations, excluding 10 ties, LIM was preferred in the majority of comparisons, indicating favorable clinical acceptance under blinded evaluation.

\subsection{Ablation Studies}

\paragraph{\textbf{Effect of Q-Sampling Trajectory Shortcut}}
Fig.~\ref{fig:ablation-t_s} evaluates different starting denoising steps across tracers. The optimal starting point varies by tracer and occurs at 5\%, 2.5\%, and 1\% of the original trajectory for $^{18}$F-FDG, $^{68}$Ga-DOTATATE, and $^{18}$F-PSMA, respectively. Interestingly, performance remains nearly unchanged over a broad range of early trajectory truncation before improving near the optimal region and eventually degrading under overly aggressive shortening. This observation suggests that the early reverse trajectory contains substantial redundancy while excessive truncation eventually removes useful refinement steps.

\begin{table}[t]
\centering
\caption{Quantitative ablation of the feature reuse interval $N$ for the proposed Feature-Reuse Network Shortcut on 10 randomly selected $^{18}$F-PSMA PET cases. To isolate the effect of feature reuse, the reverse diffusion process was initialized from the full trajectory ($t_s=T=1000$), without applying the Q-Sampling Trajectory Shortcut. Results are reported as mean $\pm$ standard deviation.}
\label{tab:ablation_feature_reuse_interval}

\renewcommand{\arraystretch}{1.1}
\setlength{\tabcolsep}{3.5pt}
\footnotesize

\begin{tabular}{lcccc}
\toprule
\footnotesize
\textbf{Interval} & \textbf{PSNR}$\uparrow$ & \textbf{SSIM}($\times10^2$)$\uparrow$ & \textbf{MAE}($\times10^{-2}$)$\downarrow$ & \textbf{Time}$\downarrow$ \\
\midrule
Baseline & 42.06$\pm$2.82 & 97.14$\pm$1.22 & 4.9862$\pm$1.0049 & 84.93 min \\
$N=3$    & 42.17$\pm$2.88 & 97.15$\pm$1.21 & 4.9632$\pm$1.0107 & 48.50 min \\
$N=5$    & 42.12$\pm$2.76 & 96.96$\pm$1.18 & 5.3839$\pm$1.0711 & 40.98 min \\
$N=10$   & 42.06$\pm$2.78 & 96.78$\pm$1.17 & 5.5847$\pm$1.0897 & 35.81 min \\
$N=25$   & 41.99$\pm$2.72 & 96.57$\pm$1.22 & 5.7681$\pm$1.1594 & 32.54 min \\
\bottomrule
\end{tabular}
\end{table}

\begin{figure}[htb!]
    \centering
    \includegraphics[width=\columnwidth]{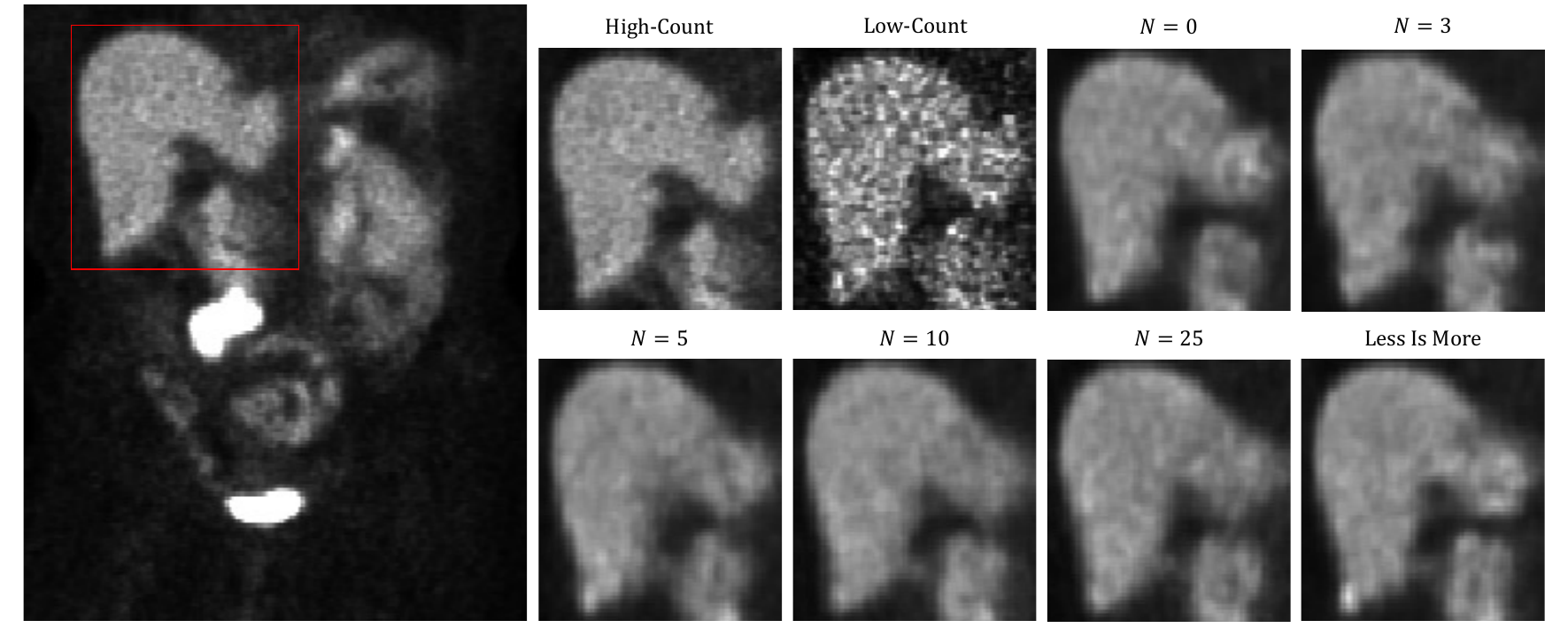}
    \vspace{-8pt}
    \caption{Visualization results under different feature reuse intervals $N$ for 10\% low-count $^{68}$Ga-DOTATATE PET denoising. A moderate reuse interval better preserves structural details and anatomical fidelity while reducing inference time, whereas overly large intervals may introduce visible degradation.}
    \label{fig: Ablation-CacheInterval}
\end{figure}

\paragraph{\textbf{Effect of Feature-Reuse Network Shortcut}}
Table~\ref{tab:ablation_feature_reuse_interval} and Fig.~\ref{fig: Ablation-CacheInterval} evaluate feature reuse independently using $t_s=T$. Moderate reuse intervals significantly reduce inference cost with minimal quality loss. In particular, $N=3$ provides the most favorable trade-off and slightly improves reconstruction metrics, suggesting that adjacent diffusion steps contain highly redundant semantic representations. Larger intervals progressively degrade image quality and introduce oversmoothing, indicating accumulated approximation error.

\begin{table}[htb!]
\centering
\caption{Quantitative robustness analysis across different low-count levels on the public $^{18}$F-FDG PET dataset. Reconstruction performance of \textbf{Standard Diffusion} and the proposed \textbf{Less Is More-Full (LIM-Full)} is compared at 25\%, 10\%, 5\%, and 2\% count levels. Results are reported as \textbf{Standard Diffusion / LIM-Full}, with higher PSNR and SSIM and lower MAE indicating better reconstruction quality.}
\label{tab:fdg_pet_normal_cache_compact}

\renewcommand{\arraystretch}{1.12}
\setlength{\tabcolsep}{5pt}
\footnotesize

\begin{tabular}{lccc}
\toprule
\textbf{$Count Level$} & \textbf{PSNR}$\uparrow$ & \textbf{SSIM}($\times 10^2$)$\uparrow$ & \textbf{MAE}($\times 10^{-2}$)$\downarrow$ \\
\midrule
25\%   & 47.06 / 48.74 & 99.04 / 99.10 & 2.5857 / 2.5019 \\
10\%  & 46.65 / 47.29 & 98.70 / 98.78 & 3.0804 / 2.9601 \\
5\%  & 45.54 / 46.08 & 98.26 / 98.36 & 3.6586 / 3.5691 \\
2\%  & 42.97 / 43.47 & 96.84 / 96.99 & 5.3196 / 5.1159 \\
\bottomrule
\label{Tab-Ablation-CountLevel}
\end{tabular}
\end{table}

\begin{figure}[htb!]
    \centering
    \includegraphics[width=0.99\columnwidth]{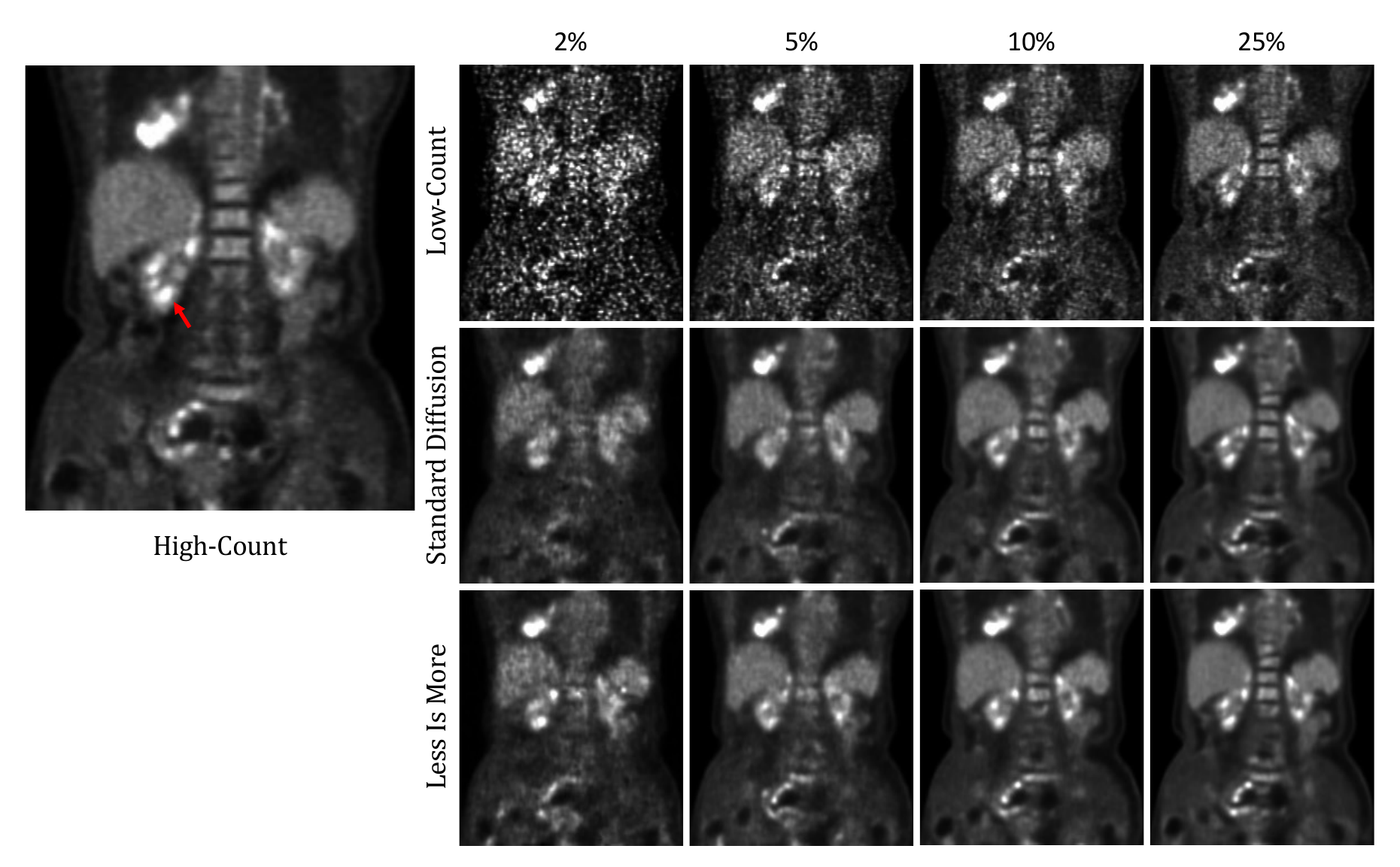}
    \vspace{-8pt}
    \caption{Visualization results across different low-count levels (25\%, 10\%, 5\%, and 2\%) for PET denoising. Representative reconstructions from Standard Diffusion and the proposed Less Is More (LIM) framework are compared against the high-count reference. While reconstruction quality gradually decreases with reduced count statistics, LIM preserves local uptake structures and anatomical details across a broad range of count levels.}
    \label{fig: ablation count level}
\end{figure}

\paragraph{\textbf{Robustness Across Count Levels}}
Finally, we evaluated LIM under multiple count levels ranging from 25\% to 2\% (Table~\ref{Tab-Ablation-CountLevel}, Fig.~\ref{fig: ablation count level}). Across all count levels, LIM consistently matched or exceeded Standard Diffusion in quantitative performance. The performance advantage becomes more noticeable under lower count conditions, suggesting that trajectory reduction remains effective even when reconstruction becomes more challenging due to reduced count statistics.

\section{Discussion}

In this study, we proposed \textit{Less Is More (LIM)}, a training-free acceleration framework for diffusion-based PET image denoising. The proposed method addresses one of the major barriers preventing practical deployment of diffusion-based restoration models in clinical imaging: although diffusion models provide strong restoration performance, iterative reverse diffusion remains computationally expensive, particularly for high-resolution volumetric PET reconstruction. By jointly introducing the proposed \emph{Q-Sampling Trajectory Shortcut} and \emph{Feature-Reuse Network Shortcut}, LIM reduces both trajectory-level and network-level redundancy during inference, enabling substantial acceleration without retraining or architectural modification.

A central observation of this work is that conditional PET restoration differs fundamentally from unconditional image generation. Existing diffusion acceleration methods are primarily developed for text-to-image and video generation, where reverse diffusion must progressively synthesize global structures and semantic content from pure Gaussian noise. In contrast, low-count PET restoration begins from an observed image that already contains substantial anatomical and tracer uptake information. Consequently, a large portion of the reverse diffusion process may become redundant because the model no longer needs to reconstruct global image structures from scratch. This observation motivates the proposed Q-Sampling Trajectory Shortcut, which initializes reverse diffusion from an intermediate state constructed from the observed low-count PET image rather than from pure noise.

The quantitative and qualitative results consistently support this hypothesis. Across both public and in-house datasets, trajectory shortening through the proposed initialization not only substantially reduced inference time but frequently maintained or improved reconstruction quality relative to the original sampling procedure. This finding suggests that early reverse diffusion steps may contribute limited practical value in conditional PET denoising. Importantly, this behavior differs from conventional scheduler replacement approaches such as DDIM, where aggressive trajectory reduction often resulted in noticeable degradation in image quality. These observations indicate that preserving compatibility with the original diffusion noise schedule while exploiting conditional structural priors is critical for effective acceleration.

Beyond global trajectory redundancy, our results further demonstrate substantial local redundancy within the denoising network itself. The proposed Feature-Reuse Network Shortcut exploits the observation that intermediate U-Net representations evolve smoothly across neighboring denoising steps. By caching and reusing temporally similar features, LIM further reduced computational cost while preserving reconstruction fidelity. Interestingly, moderate feature reuse occasionally produced slight improvements in quantitative metrics. One possible explanation is that feature reuse behaves as a weak regularization mechanism by suppressing unstable high-frequency perturbations introduced through repeated iterative updates while preserving dominant anatomical structures and tracer uptake patterns. Although this interpretation remains speculative, it suggests an interesting direction for future work toward understanding feature evolution and information redundancy in diffusion-based medical image restoration.

Another important finding is the robustness of the proposed framework across multiple tracers and count levels. Although the pretrained diffusion model was originally developed using public $^{18}$F-FDG data, LIM generalized favorably to unseen tracer distributions, including $^{68}$Ga-DOTATATE and $^{18}$F-PSMA PET, without retraining. Furthermore, experiments under varying count levels demonstrated that the acceleration strategy remained effective under increasingly challenging acquisition conditions. These findings suggest that the redundancy exploited by LIM is not tracer-specific but may reflect broader properties of conditional diffusion restoration.

From a clinical perspective, accelerating diffusion-based PET denoising may improve the practicality of low-count PET protocols. Reduced-count imaging has the potential to lower radiation exposure, shorten acquisition duration, improve scanner throughput, and enhance patient comfort. However, such gains are only meaningful if denoising methods preserve clinically relevant information. Beyond conventional image-level metrics, our blinded reader study showed that LIM achieved improved clinical confidence and lesion visibility while maintaining comparable noise suppression relative to Standard Diffusion. Interestingly, Standard Diffusion occasionally produced smoother images but also tended to suppress lesion conspicuity, suggesting that lower visual noise does not necessarily translate into improved perceived diagnostic utility.

Despite these encouraging results, several limitations should be acknowledged. First, the optimal starting timestep $t_s$ and feature reuse interval $N$ were selected empirically and may vary across tracers, scanners, reconstruction protocols, and pretrained diffusion models. Future work may investigate adaptive acceleration policies guided by uncertainty estimation or trajectory dynamics. Second, evaluation was primarily based on image-level similarity metrics and reader preference. More PET-specific analyses, including SUV bias, contrast recovery, lesion detectability, and downstream clinical tasks, should be incorporated in future studies. Third, although we validated the framework across multiple tracers and a public external dataset, broader multi-center evaluation remains necessary to assess robustness across vendors and acquisition settings.

Overall, our findings suggest that diffusion-based PET restoration contains substantial trajectory-level and feature-level redundancy that can be exploited to dramatically improve inference efficiency without sacrificing reconstruction quality. Beyond PET denoising, the proposed Global--Local Skipping paradigm may provide a broader perspective for developing efficient diffusion inference algorithms for conditional medical image restoration.

\section{Conclusion}
In this work, we presented \textit{Less Is More}, a practical training-free acceleration framework for diffusion-based 3D PET image denoising that combines the proposed Q-Sampling Trajectory Shortcut and Feature-Reuse Network Shortcut to reduce both reverse trajectory length and redundant network computation. Extensive experiments across multiple PET tracers ($^{18}$F-FDG, $^{68}$Ga-DOTATATE, and $^{18}$F-PSMA), together with external validation on a public PET dataset and blinded reader evaluation, demonstrated that the proposed method substantially improves inference efficiency while maintaining and in some cases improving reconstruction fidelity and perceived clinical utility. More importantly, our findings suggest that the full reverse diffusion trajectory may contain considerable redundancy in conditional medical image restoration tasks, where degraded inputs already provide strong anatomical and functional priors. These observations provide a restoration-oriented perspective on efficient diffusion inference and highlight opportunities for developing adaptive trajectory control and feature reuse strategies toward practical deployment of diffusion-based restoration models in real-world 3D medical imaging applications.

\section{Acknowledgments}
Parts of the data used in preparation of this article were obtained from the University of Bern, Dept. of Nuclear Medicine and School of Medicine, Ruijin Hospital. As such, the investigators contributed to the design and implementation of DATA and/or provided data but did not participate in analysis or writing of this report. A complete listing of investigators can be found at: “https://udpet-challenge.github.io/”


\bibliographystyle{model2-names.bst}\biboptions{authoryear}
\bibliography{refs}


\newpage
\appendix
\section{LIM Rational for PET Imaging}
\label{Appendix}

\subsection{Low-count PET and high-count PET share the same underlying structures}

Positron Emission Tomography (PET) records coincidence events arising from pairs of 511 keV photons generated by positron--electron annihilation. In this study, as described in Sec.~\ref{sec:Baselines_And_Eval_Strategies}, we used the DRF=10 setting for both our in-house tracer-specific datasets and the publicly available UDPET Challenge 2025 dataset, corresponding to PET images reconstructed from approximately 10\% of the original list-mode acquisition data.

The down-sampled list-mode data were reconstructed using the ordered-subsets expectation maximization (OSEM) algorithm  to generate low-count PET images. The corresponding high-count PET images were reconstructed from the original list-mode data without down-sampling using the same reconstruction protocol and served as high-count reference images. Because each low-count PET image and its high-count reference originate from the same list-mode acquisition, they are expected to preserve the same underlying tracer distribution and anatomical structures. The primary difference between them arises from the reduced number of detected coincidence events used for reconstruction, which increases statistical noise and degrades image quality in the low-count PET image. Therefore, the low-count PET image can be regarded as a noisy observation of the same anatomical and functional structures present in the high-count PET image, rather than an image with different underlying content.

Although reducing the number of PET events inevitably discards information and increases reconstruction uncertainty, the low-count image still retains most of the global anatomical layout, tracer uptake distribution, and lesion-related structural information. Therefore, we model the low-count and high-count PET images as noisy observations of the same underlying signal, but with different noise levels.

Formally, let \(x^{\mathrm{HC}}(r)\) and \(x^{\mathrm{LC}}(r)\) denote the high-count and low-count PET image intensities at voxel location \(r \in \Omega\), respectively. Under the approximation that both images are reconstructed from the same underlying tracer distribution, we assume
\begin{equation}
    \mathbb{E}\big[x^{\mathrm{HC}}(r)\big]
    \approx
    \mathbb{E}\big[x^{\mathrm{LC}}(r)\big]
    =
    \mu(r),
\end{equation}
where \(\mu(r)\) denotes the underlying expected PET signal at voxel \(r\). The reduced number of detected events increases the uncertainty of the reconstructed image, leading to a higher noise level in the low-count image:
\begin{equation}
    \mathrm{Var}\big[x^{\mathrm{LC}}(r)\big]
    \ge
    \mathrm{Var}\big[x^{\mathrm{HC}}(r)\big].
\end{equation}

Equivalently, by viewing the reconstructed PET images as random vectors in \(\mathbb{R}^{V}\) over \(V\) voxels, we write
\begin{equation}
    \mathbb{E}\big[x^{\mathrm{HC}}\big]
    \approx
    \mathbb{E}\big[x^{\mathrm{LC}}\big]
    =
    \mu,
\end{equation}
and characterize the higher noise level of the low-count image by its larger expected deviation from the underlying signal:
\begin{equation}
    \mathbb{E}\left[\left\|x^{\mathrm{LC}}-\mu\right\|_2^2\right]
    \ge
    \mathbb{E}\left[\left\|x^{\mathrm{HC}}-\mu\right\|_2^2\right].
\end{equation}

This assumption can be intuitively understood from the list-mode acquisition process. Let \(\{Y_i\}_{i=1}^{N}\) denote the detected list-mode events from a full acquisition. Ignoring reconstruction nonlinearity for the purpose of statistical intuition, the reconstructed high-count image intensity at voxel \(r\) can be idealized as an aggregation of event contributions:
\begin{equation}
    x^{\mathrm{HC}}(r)
    =
    \frac{1}{N}\sum_{i=1}^{N} f_r(Y_i),
\end{equation}
where \(f_r(Y_i)\) denotes the contribution of event \(Y_i\) to voxel \(r\). A low-count image reconstructed from a reduced subset of list-mode events \(S \subset \{1,\dots,N\}\), with \(|S| \approx 0.1N\), can be similarly idealized as
\begin{equation}
    x^{\mathrm{LC}}(r)
    =
    \frac{1}{|S|}\sum_{i\in S} f_r(Y_i).
\end{equation}
Since \(x^{\mathrm{LC}}(r)\) is estimated from fewer detected events, it has a higher sampling variance than \(x^{\mathrm{HC}}(r)\), while still estimating approximately the same underlying signal \(\mu(r)\). This supports our central modeling assumption: low-count and high-count PET images share the same anatomical and functional structures in expectation, whereas the low-count image is primarily degraded by increased statistical noise and reduced perceptual image quality.

\subsection{Q-sampled low-count PET as an approximate intermediate initialization}

We next consider the DDPM forward noising process applied to the low-count PET image. 
Given a denoising timestep \(t\), the standard DDPM forward process defines the marginal distribution
\begin{equation}
    q(\mathbf{x}_t \mid \mathbf{x}_0)
    =
    \mathcal{N}\!\Big(
        \sqrt{\bar{\alpha}_t}\,\mathbf{x}_0,\,
        (1-\bar{\alpha}_t)\mathbf{I}
    \Big),
\end{equation}
where \(\bar{\alpha}_t = \prod_{s=1}^{t}\alpha_s\). Equivalently, a noisy sample at timestep \(t\) can be obtained by
\begin{equation}
    \mathbf{x}_t
    =
    \sqrt{\bar{\alpha}_t}\,\mathbf{x}_0
    +
    \sqrt{1-\bar{\alpha}_t}\,\boldsymbol{\varepsilon},
    \qquad
    \boldsymbol{\varepsilon}\sim\mathcal{N}(\mathbf{0},\mathbf{I}).
\end{equation}

In the proposed Q-sampling Trajectory Shortcut, we replace the unknown high-count target \(\mathbf{x}^{\mathrm{HC}}\) with the observed low-count image \(\mathbf{x}^{\mathrm{LC}}\), and construct the intermediate initialization at the selected starting timestep \(t_s\) as
\begin{equation}
    \mathbf{x}_{t_s}
    =
    \sqrt{\bar{\alpha}_{t_s}}\,\mathbf{x}^{\mathrm{LC}}
    +
    \sqrt{1-\bar{\alpha}_{t_s}}\,\boldsymbol{\varepsilon},
    \qquad
    \boldsymbol{\varepsilon}\sim\mathcal{N}(\mathbf{0},\mathbf{I}).
\end{equation}

This initialization can be interpreted as an approximate sample from the intermediate diffusion state that would have been obtained by applying the forward process to the high-count PET image:
\begin{equation}
    q(\mathbf{x}_{t_s}\mid \mathbf{x}^{\mathrm{HC}})
    =
    \mathcal{N}\!\Big(
        \sqrt{\bar{\alpha}_{t_s}}\,\mathbf{x}^{\mathrm{HC}},
        (1-\bar{\alpha}_{t_s})\mathbf{I}
    \Big).
\end{equation}
Since \(\mathbf{x}^{\mathrm{LC}}\) and \(\mathbf{x}^{\mathrm{HC}}\) share approximately the same underlying anatomical and tracer uptake structures, replacing \(\mathbf{x}^{\mathrm{HC}}\) with \(\mathbf{x}^{\mathrm{LC}}\) mainly introduces an additional low-count noise component while preserving the global image content. Therefore, \(\mathbf{x}_{t_s}\) provides a structurally informed approximation to the desired intermediate diffusion state.

This design has two important advantages. First, the initialization retains task-relevant information from the low-count PET image, including global anatomical layout, tracer distribution, and lesion-related uptake patterns. Second, the added Gaussian perturbation matches the noise scale expected by the pre-trained diffusion model at timestep \(t_s\). Therefore, the reverse process can start from an intermediate state that is both informative and compatible with the learned denoising trajectory, rather than from an unstructured Gaussian sample at timestep \(T\).

\subsection{Why starting from \(t_s\) can be better than full reverse sampling}

Although the standard diffusion inference procedure starts from pure Gaussian noise \(\mathbf{x}_T \sim \mathcal{N}(\mathbf{0},\mathbf{I})\), this strategy may not be optimal for paired PET denoising. In image generation tasks, sampling from \(\mathbf{x}_T\) is necessary because no image-specific observation is available. In contrast, PET denoising is a conditional restoration problem: as we observed from Fig.~\ref{Fig-Intro}, the low-count PET image is already available and contains substantial structural information about the desired high-count image. Therefore, starting the reverse process from pure Gaussian noise may introduce unnecessary stochasticity and require the model to recover image structures that are already present in the low-count input.

A possible explanation is that the learned reverse process may accumulate approximation errors over a long denoising trajectory. Let \(p_\theta(\mathbf{x}_{t-1}\mid \mathbf{x}_t)\) denote the learned reverse transition and \(p^\star(\mathbf{x}_{t-1}\mid \mathbf{x}_t)\) denote the ideal reverse kernel. Although each transition may be locally accurate,
\begin{equation}
    p_\theta(\mathbf{x}_{t-1}\mid \mathbf{x}_t)
    \approx
    p^\star(\mathbf{x}_{t-1}\mid \mathbf{x}_t),
\end{equation}
small discrepancies can accumulate when the model repeatedly composes many reverse transitions from \(T\) to \(0\). This accumulated error may be particularly relevant in 3D PET restoration, where the model must preserve subtle anatomical boundaries, tracer uptake patterns, and lesion-related details over a long reverse trajectory.

By initializing the reverse process from a Q-sampled low-count PET image at timestep \(t_s\), the number of reverse transitions is reduced from \(T\) to \(t_s\). This shorter trajectory can reduce cumulative approximation error and inference cost. Meanwhile, because the initialization is constructed from the low-count image, it preserves global structures that are already close to the desired high-count image in expectation. As a result, the model is asked to solve a simpler restoration problem: refining and denoising a structurally meaningful intermediate state rather than synthesizing the PET image from pure Gaussian noise.

However, the starting timestep \(t_s\) must be properly selected. If \(t_s\) is too large, excessive Gaussian perturbation may destroy useful structural information in the low-count image, weakening the benefit of image-specific initialization. If \(t_s\) is too small, the reverse trajectory becomes very short, but the model may not have enough denoising capacity to correct low-count noise and reconstruction artifacts. Therefore, \(t_s\) controls a trade-off between structural preservation, denoising capacity, and computational efficiency.

This explains why an intermediate starting timestep can outperform full reverse sampling in our experiments: it avoids unnecessary early reverse denoising steps from pure noise, reduces cumulative denoising error, and leverages the low-count PET image as an informative initialization while maintaining compatibility with the diffusion noise schedule.

\subsection{Ablation studies on noise-level matching}

To further validate the mechanism of the proposed Q-sampling trajectory shortcut, we conducted an ablation study by decoupling the reverse starting timestep from the noise level used for initialization. In the proposed method, the low-count PET image is noised according to the same timestep from which reverse denoising starts. That is, when the reverse process starts from timestep \(t_s\), the initialization is generated using the forward noising level \(\bar{\alpha}_{t_s}\):
\begin{equation}
    \mathbf{x}_{t_s}^{\mathrm{init}}
    =
    \sqrt{\bar{\alpha}_{t_s}}\,\mathbf{x}^{\mathrm{LC}}
    +
    \sqrt{1-\bar{\alpha}_{t_s}}\,\boldsymbol{\varepsilon}.
\end{equation}

To test whether this timestep--noise-level matching is necessary, we manually changed the noise level used for initialization while keeping the reverse starting timestep fixed. Specifically, for a reverse process starting from timestep \(t\), we generated the initial state using three different noise levels: \(2t\), \(t\), and \(0.5t\). The \(t\) setting corresponds to the proposed matched initialization, whereas \(2t\) and \(0.5t\) intentionally introduce a mismatch between the noise level of the input and the denoising timestep expected by the model.

\begin{table}[htb!]
\centering
\caption{Ablation on timestep--noise-level matching. Each entry reports PSNR / SSIM.}
\label{Tab-AblationStudy}

\renewcommand{\arraystretch}{1.10}
\setlength{\tabcolsep}{3.5pt}
\footnotesize

\begin{tabular}{@{}c c c c@{}}
\toprule
\multirow{2}{*}{\textbf{Start}} &
\multicolumn{3}{c}{\textbf{Initialization noise level}} \\
\cmidrule(lr){2-4}
& \textbf{\(2t\)} & \textbf{\(t\)} & \textbf{\(0.5t\)} \\
\midrule
\(t=100\) & 22.91/21.78 & \textbf{44.33/98.47} & 43.51/98.13 \\
\(t=50\)  & 26.53/17.43 & \textbf{46.66/98.56} & 44.73/98.20 \\
\(t=10\)  & 36.47/70.48 & \textbf{46.52/98.77} & 46.09/98.41 \\
\bottomrule
\end{tabular}
\end{table}

As shown in Table~\ref{Tab-AblationStudy}, the matched setting consistently achieves the best performance across different starting timesteps. When the initialization noise level is larger than the reverse starting timestep, such as using the \(2t\) noise level while starting reverse denoising from timestep \(t\), the input is noisier than what the model expects at that timestep. This mismatch severely degrades image quality, as reflected by the large performance drop in the \(2t\) column. In this case, the reverse process starts from an over-corrupted image but is given too few denoising steps to recover the lost structures.

Conversely, when the initialization noise level is smaller than the reverse starting timestep, such as using the \(0.5t\) noise level, the input contains less Gaussian noise than expected by the model at timestep \(t\). Although this setting preserves more low-count image structure, it introduces a distribution mismatch between the actual input and the timestep condition used by the denoising network. As a result, the model applies a denoising operation calibrated for a noisier state, which can lead to suboptimal restoration.

These results support the key design of Q-sampling trajectory shortcut: the reverse starting timestep and the forward noising level should be coupled. The low-count PET image provides image-specific structural information, while the Gaussian perturbation aligns the initialization with the diffusion model's expected noise distribution at timestep \(t_s\). Therefore, the proposed matched Q-sampling initialization provides a more reliable intermediate starting point than mismatched noise-level initialization.

\end{document}